\documentclass[runningheads]{llncs}

\usepackage{eccv}

\usepackage{eccvabbrv}

\usepackage{graphicx}
\usepackage{array}         
\usepackage{multirow}      
\usepackage{booktabs}      
\usepackage{colortbl}      
\usepackage{tabularx}      
\usepackage{makecell}      
\usepackage{bm}            
\usepackage{pifont}        
\usepackage{wrapfig}
\usepackage{float}

\newcommand{\cmark}{\textcolor{GreenColor}{\ding{51}}\xspace}
\newcommand{\xmark}{\ding{55}\xspace}

\definecolor{PurpleColor}{HTML}{8B008B}
\definecolor{OrangeColor}{rgb}{0.914,0.541,0.141}
\definecolor{GreenColor}{rgb}{0.137,0.573,0.565}

\definecolor{ttt3r}{HTML}{DFA7A7}
\definecolor{cut3r}{HTML}{B8A5C7}
\definecolor{gt}{HTML}{C9C9C9}

\definecolor{yzybest}{rgb}{0.98, 0.8, 0.8}
\definecolor{yzysecond}{rgb}{0.99, 0.88, 0.77}
\definecolor{yzythird}{rgb}{1.0, 1.0, 0.8}

\newcommand{\firstc}{\cellcolor{yzybest}}
\newcommand{\secondc}{\cellcolor{yzysecond}}

\newcommand{\cg}{\color{gray!60}}

\usepackage{hyperref}

\usepackage{orcidlink}

\begin{document}

\title{RayMap3R: Inference-Time RayMap for Dynamic 3D Reconstruction}

\titlerunning{RayMap3R}

\author{Feiran Wang\inst{1} \and
Zezhou Shang\inst{1} \and
Gaowen Liu\inst{2} \and
Yan Yan$^{1,\dagger}$}

\authorrunning{F.~Wang et al.}

\institute{$^1$ University of Illinois Chicago \qquad $^2$ Cisco Research \\[2pt] {\small $^\dagger$ Corresponding author}}

\maketitle

\vspace{-1em}
\begin{figure}[H]
    \centering
    \includegraphics[width=\textwidth]{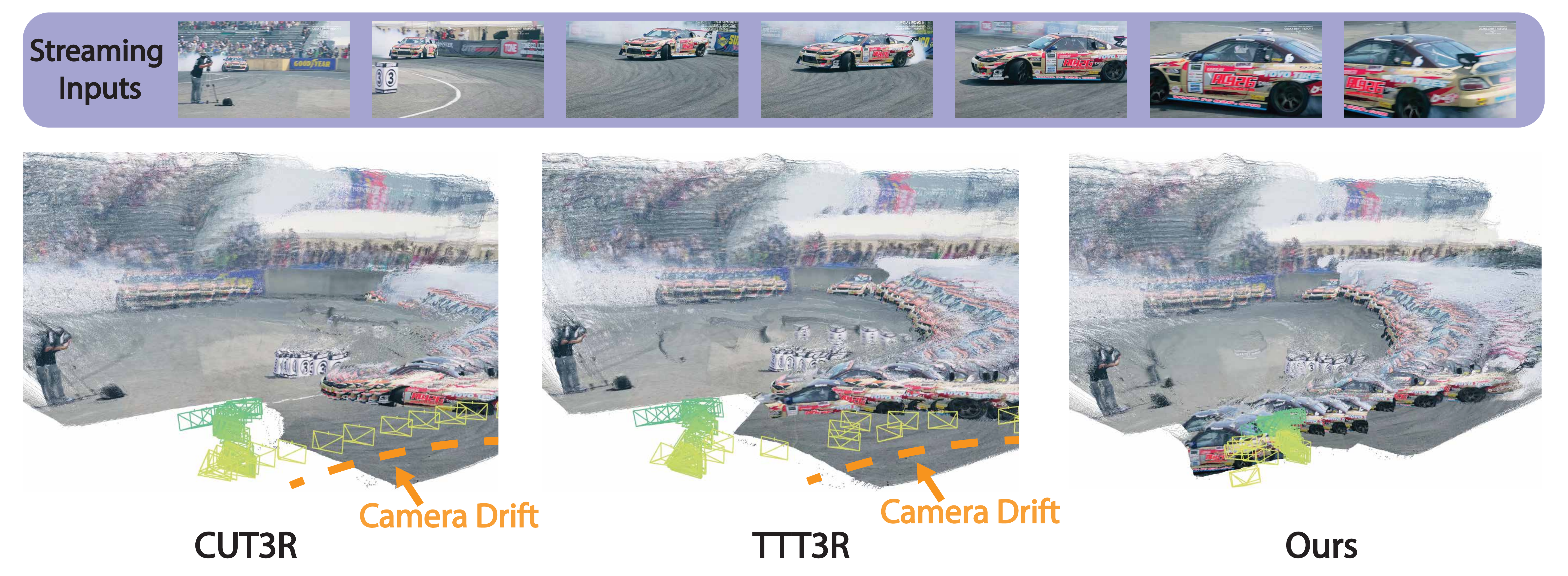}
    \vspace{-4ex}
    \caption{\textbf{Streaming 3D Reconstruction for Dynamic Scenes.}
    RayMap3R leverages the static bias of RayMap-only predictions to identify and suppress dynamic regions at inference time, reducing camera drift and improving geometry fidelity.
    }
    \label{fig:teaser}
    \vspace{-1em}
\end{figure}

\begin{abstract}


Streaming feed-forward 3D reconstruction enables real-time joint estimation of scene geometry and camera poses from RGB images. However, without explicit dynamic reasoning, streaming models can be affected by moving objects, causing artifacts and drift. In this work, we propose RayMap3R, a training-free streaming framework for dynamic scene reconstruction. 
We observe that RayMap-based predictions exhibit a static-scene bias, providing an internal cue for dynamic identification. Based on this observation, we construct a dual-branch inference scheme that identifies dynamic regions by contrasting RayMap and image predictions, suppressing their interference during memory updates.
We further introduce reset metric alignment and state-aware smoothing to preserve metric consistency and stabilize predicted trajectories. Our method achieves state-of-the-art performance among streaming approaches on dynamic scene reconstruction across multiple benchmarks.
The project page and code are available at \url{https://raymap3r.github.io/}.

\end{abstract}

\section{Introduction}
\label{sec:intro}

Feed-forward 3D models have achieved remarkable progress in
reconstructing 3D structures such as point clouds, depth maps, and
camera poses from monocular images.
An influential class of
methods~\cite{wang2024dust3r, monst3r, wang2025vggt, duisterhof2024mast3r, vggtlong, zhuo2025streaming}
adopts an offline reconstruction paradigm that builds pairwise
correspondences between all input images, forming dense attention maps
that enable precise estimation of geometry and camera poses.
However, the offline paradigm requires the complete set of frames
before producing output, making it unsuitable for real-time
applications.
Moreover, both computational and memory costs scale rapidly with
sequence length, for instance requiring up to 48 GB for sequences of only 200 frames~\cite{vggtslam}.

To extend feed-forward reconstruction to real-time processing, recent
works~\cite{spann3r, cut3r, ttt3r, point3r} focus on streaming
feed-forward 3D reconstruction, which performs continuous inference
from input image streams while maintaining stable computational
resources.
These methods introduce memory mechanisms that interact directly with 
incoming images, enabling high-frequency real-time inference with 
constant memory usage even over thousands of frames. 
Concretely, approaches include spatial memory banks, implicit recurrent states, and explicit point anchors, offering different trade-offs between efficiency and reconstruction fidelity.

However, streaming feed-forward 3D reconstruction models still face major limitations.
First, these models lack explicit mechanisms to
identify dynamic objects, partly due to the scarcity of training data with dynamic annotations.
Unlike offline methods that leverage dense pairwise matching across
all frames, streaming methods process frames sequentially with limited
historical context, making them particularly vulnerable to dynamic
interference.
Second, frame-by-frame prediction may introduce pose noise that accumulates over long sequences.

Existing approaches to dynamic scene
reconstruction~\cite{monst3r, li2024megasam, chen2025back, team2025aether} typically require external modules
such as flow estimators, adding overhead and domain-specific
dependencies.
RayMap~\cite{zhang2024cameras}, a per-pixel camera-ray representation,
has been widely adopted in recent streaming
models~\cite{cut3r, wang2025vggt, mapanything} to strengthen
the link between appearance and geometry.
We observe that models trained with RayMap tend to exhibit a static-scene bias, which can be exploited to identify dynamic regions at inference time without additional training or external models.

Specifically, when queried with only RayMap features, the model prioritizes static structure while suppressing dynamic regions. This bias arises because RayMap tokens encode only camera geometry, forcing the model to reconstruct scene content solely from memory. Since static structures are observed consistently across frames, they are recalled reliably, whereas dynamic objects appear transiently and tend to be suppressed. The discrepancy between RayMap-only and image-based predictions thus provides a signal exploitable for training-free dynamic identification. We provide further analysis of this property in \cref{ssec:observation}.

Building on this observation, we propose \textbf{RayMap3R}, an inference-time framework for dynamic scene reconstruction. We employ a dual-branch inference scheme: the main branch processes both image and RayMap features for geometry estimation, while the RayMap branch uses only pose-derived RayMap features to produce static-biased predictions. By comparing these predictions, we derive staticness weights that modulate memory state updates, suppressing interference from dynamic regions while preserving static structure. To preserve metric consistency across memory reset boundaries, we introduce reset metric alignment, estimating corrective transformations from repeated frames to restore alignment between segments. We further propose state-aware smoothing to stabilize predicted trajectories by using the magnitude of internal state changes as an uncertainty signal to adaptively smooth pose predictions online.

We evaluate RayMap3R on benchmarks covering camera pose estimation, depth prediction, and 3D reconstruction across both synthetic and real-world datasets. RayMap3R achieves state-of-the-art performance among streaming methods, with particularly strong gains on dynamic scenes under both per-sequence and metric-scale settings, while producing high-quality 3D reconstructions at real-time speeds with constant memory usage.

Our key contributions are:
\begin{itemize}
\item 
We observe that RayMap predictions exhibit a static-scene bias, and exploit this bias to derive staticness weights for dynamic-aware memory updates.
\item We introduce reset metric alignment and state-aware
smoothing to stabilize trajectory estimation across long sequences.
\item Extensive experiments show that RayMap3R achieves leading performance among streaming methods across benchmarks with real-time efficiency.
\end{itemize}


\section{Related Work}
\label{sec:relatedwork}

\noindent\textbf{Offline Reconstruction Models.}
Offline reconstruction methods require the complete set of input
images before producing
outputs~\cite{wang2024dust3r,lu2024align3r, monst3r, vggtlong,wang2025vggt},
employing global optimization or full-sequence attention to prioritize
accuracy over real-time performance.
DUSt3R~\cite{wang2024dust3r} pioneered pointmap representation for
scene-level 3D reconstruction, inferring camera poses and aligned
point clouds from image pairs.
Subsequent approaches~\cite{lu2024align3r, monst3r, sucar2025dynamic, duisterhof2024mast3r}
extended this framework but require pairwise processing with time-consuming optimization.
MASt3R~\cite{duisterhof2024mast3r} augments DUSt3R with dense local
features, while Fast3R~\cite{fast3r} achieves reconstruction in a
single forward.
VGGT~\cite{wang2025vggt} employs a DPT backbone for joint pose and
geometry estimation, achieving highly accurate results, though
computational cost increases rapidly with input count, limiting
scalability to long sequences. VGGT-Long~\cite{vggtlong} introduces
chunk-based reconstruction, but memory grows with chunk size,
restricting real-time applicability.

\noindent\textbf{Streaming Reconstruction Models.}
To enable real-time 3D reconstruction, recent works introduce memory
mechanisms for continuous
inference~\cite{spann3r, zhuo2025streaming, point3r,cut3r,ttt3r}.
Spann3R~\cite{spann3r} extends DUSt3R with external spatial memory
that retains relevant 3D information across frames, achieving fast
reconstruction with efficient memory usage but remaining less robust in dynamic scenes.
StreamVGGT~\cite{zhuo2025streaming} distills VGGT and introduces a
spatial-temporal decoder for streaming prediction, but memory
consumption grows with frame count.
Point3R~\cite{point3r} maintains explicit 3D point anchors to
preserve and match historical information.
CUT3R~\cite{cut3r} adopts a recurrent design with implicit memory
cache and location dictionary for efficient lookup, achieving
continuous real-time reconstruction.
TTT3R~\cite{ttt3r} extends CUT3R with a soft gating mechanism based
on cross-attention to mitigate memory forgetting, enabling stable
long-range reconstruction.
However, without explicit dynamic supervision during training,
streaming models remain susceptible to interference from
moving objects.

\noindent\textbf{Structure-from-Motion and Visual SLAM.}
Traditional SLAM
systems~\cite{agarwal2011building, campos2021orb, schonberger2016structure, davison2007monoslam, pollefeys2008detailed, mur2015orb}
rely on feature correspondence and bundle adjustment for camera pose
estimation but frequently fail in low-texture or moving object
scenarios.
Learning-based methods like DROID-SLAM~\cite{teed2021droid} incorporate
differentiable optimization frameworks, though they remain vulnerable
to dynamic content.
AnyCam~\cite{wimbauer2025anycam} fuses depth with optical flow for
accurate pose estimation; DPVO~\cite{teed2023deep} leverages
patch-level features;
MASt3R-GA~\cite{duisterhof2024mast3r} integrates learned features.
VGGT-SLAM~\cite{vggtslam} augments VGGT with SL4-based backend
refinement, enhancing trajectory consistency and reconstruction
quality over extended sequences.
However, these methods either require iterative optimization or
additional motion estimation modules, incurring substantial overhead.

\noindent\textbf{Dynamic Scene Reconstruction.}
Existing approaches generally fall into three categories:
flow-based methods~\cite{monst3r, li2024megasam} rely on
external flow estimators with hand-tuned thresholds that generalize
poorly; segmentation-based methods~\cite{team2025aether,
ravi2024sam} isolate dynamic objects via learned models yet remain
constrained by training categories; and tracking-based
methods~\cite{chen2025back} require known camera intrinsics and
iterative optimization. All three introduce additional modules with
substantial overhead, limiting generalization. In contrast, our work
reveals that models trained with the RayMap paradigm tend to exhibit a
static-scene bias exploitable for training-free dynamic
reconstruction without additional supervision.

\section{Methods}
\label{sec:methods}

Our method takes a stream of input images and predicts
camera poses, depths, and point clouds.
In \cref{ssec:raymap}, we introduce RayMap and the memory mechanism
for streaming reconstruction.
In \cref{ssec:observation}, we show that RayMap-based predictions
exhibit a static-scene bias useful for dynamic identification.
In \cref{ssec:remapping}, we propose a dual-branch inference scheme
that leverages this bias to identify dynamic regions by contrasting
image-based and RayMap-only predictions.
Finally, in \cref{ssec:reset,ssec:smoothing}, we introduce
reset metric alignment and state-aware smoothing to stabilize
trajectory estimation over long sequences.

\subsection{RayMap and Memory Mechanism}
\label{ssec:raymap}
\noindent\textbf{RayMap Representation.}
RayMap is an $H\times W\times 6$ per-pixel tensor encoding the
\emph{ray origin} and \emph{unit direction} for each pixel, determined
by camera intrinsics $K\in\mathbb{R}^{3\times3}$ and extrinsics
$\mathbf{T}=[R\,|\,\boldsymbol{\tau}]$~\cite{zhang2024cameras, cut3r, gao2024cat3d}.
For a pixel in homogeneous coordinates
$\tilde{\mathbf{u}}=(x,y,1)^\top$, the RayMap is defined as
\begin{equation}
\text{RayMap}(x,y)=\big[\,\mathbf{c},\,\widehat{\mathbf{d}}(x,y)\,\big]\in\mathbb{R}^{6},
\end{equation}
where the ray origin and unit direction in world coordinates:
\begin{equation}
\mathbf{c}=-R^{\top}\boldsymbol{\tau}, \qquad
\widehat{\mathbf{d}}(x,y)=\frac{R^{\top}K^{-1}\tilde{\mathbf{u}}}{\|R^{\top}K^{-1}\tilde{\mathbf{u}}\|_2}.
\end{equation}
Under the pinhole model, $\mathbf{c}$ remains constant across all
pixels as it represents the camera center, while
$\widehat{\mathbf{d}}(x,y)$ encodes the
viewing direction of each ray.

\noindent \textbf{Memory Mechanism.}
Streaming 3D reconstruction models maintain a memory cache for
continuous interaction with images.
Methods such as CUT3R~\cite{cut3r} and TTT3R~\cite{ttt3r} employ an
implicit memory cache, enabling stable memory usage and fast inference
over thousands of frames.
The implicit memory is represented as a latent state $s_t$, encoding
the understanding of the 3D scene at timestep $t$.

\noindent\textbf{Training and Inference.}
During training, input image $I_t$ is patchified into image tokens
$f_t$, while the corresponding camera pose is transformed into RayMap
and patchified into RayMap tokens $r_t$. These tokens interact with
the state through two strategies: either image tokens $f_t$ combined
with RayMap tokens $r_t$, or RayMap tokens $r_t$ alone. The state
updates from $s_{t-1}$ to $s_t$ by integrating information from $f_t$
and $r_t$, then the query process decodes $s_t$ to predict camera
pose, depth, and point clouds. During inference, after a short warmup
period, the model can predict 3D information from either images or
RayMap alone, enabling geometry prediction from arbitrary viewpoints
without image input.

\subsection{Static Bias of RayMap Predictions}
\label{ssec:observation}
We observe that streaming 3D reconstruction models trained with
the RayMap paradigm~\cite{cut3r, ttt3r} exhibit a static scene
bias when making predictions from RayMap tokens alone. As described in
\cref{ssec:raymap}, the model can predict geometry using only RayMap tokens
$r_t$ and the scene representation in $s_{t-1}$. Without appearance
information from image features $f_t$, the model relies solely on geometric
consistency in $r_t$, and tends to prioritize static structure over
dynamic content.

\begin{figure}[t]
    \centering
    \includegraphics[width=1.0\linewidth]{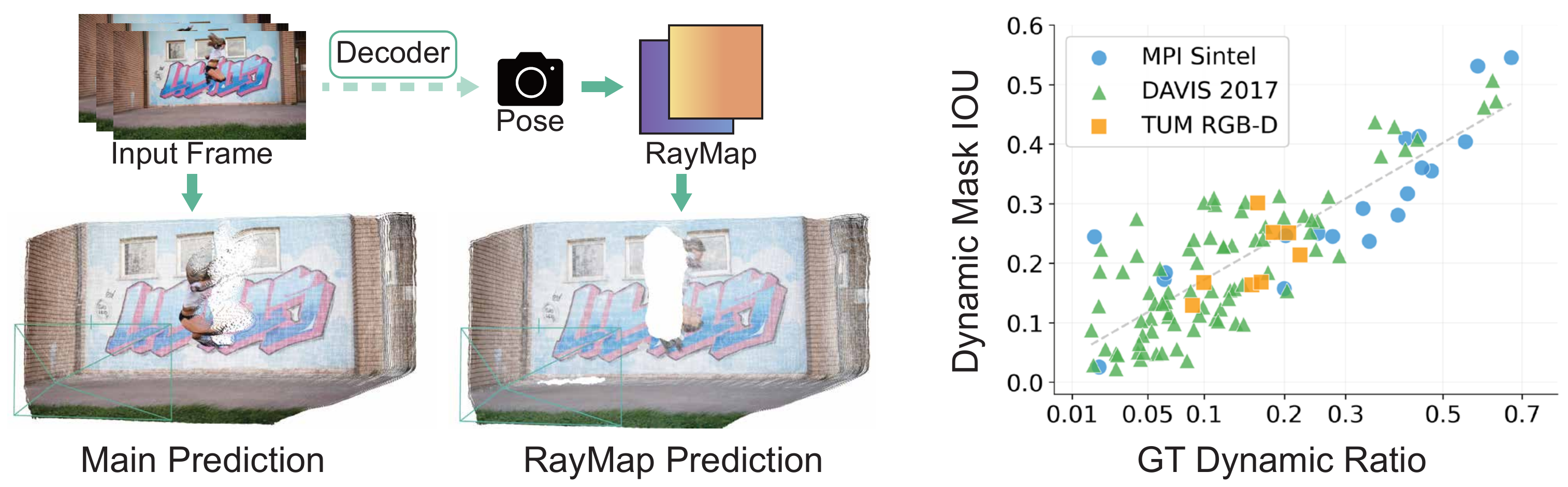}
    \caption{
    \textbf{RayMap Static Bias.} \textbf{Left:} 
    The main prediction and RayMap prediction produce differing depth estimates in dynamic regions, suggesting a static bias in RayMap predictions.
    \textbf{Right:} 
    The depth discrepancy correlates with ground-truth dynamic ratio across multiple datasets, suggesting a consistent signal for dynamic identification.
    }
    \label{fig:static_raymap}
    \vspace{-0.5em}
\end{figure}

As illustrated in \cref{fig:static_raymap} (left), given the same memory state
$s_{t-1}$, the image-based main prediction reconstructs the scene including the
dynamic foreground, while the RayMap-based prediction from the same camera pose
produces geometry focused on static background, suppressing the influence of
moving objects. We attribute this bias to two factors. First, training datasets
are dominated by static scenes with limited dynamic annotations, causing models
to develop a prior toward static structures. Second, RayMap tokens encode only
camera geometry without appearance cues, leading the model to rely on
temporally consistent structure stored in memory rather than frame-specific
dynamic content.

\begin{figure}[t]
    \centering
    \includegraphics[width=\linewidth]{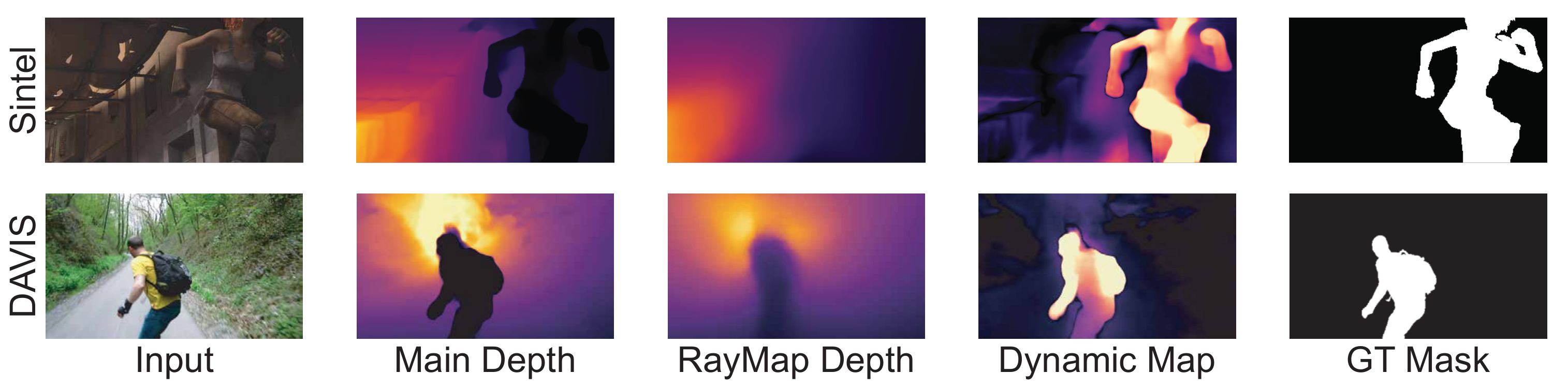}
    \caption{
       \textbf{Dynamic Map Visualization.} The dynamic map is the per-pixel depth difference between the main branch and RayMap-only predictions, closely aligning with ground-truth dynamic masks across both synthetic and real scenes.
    }
    \label{fig:raymap_compare}

    \vspace{-1.5em}
\end{figure}

To examine whether this property holds generally, we compute the per-pixel
depth discrepancy between the two branches across 108 sequences from MPI
Sintel~\cite{butler2012naturalistic}, DAVIS 2017~\cite{perazzi2016benchmark},
and TUM RGB-D~\cite{sturm2012benchmark}, and measure its overlap with ground-truth dynamic masks via IoU. As shown in
\cref{fig:static_raymap} (right), the dynamic mask IoU correlates positively with the ground-truth dynamic ratio (Spearman $\rho = 0.77$,
$p < 10^{-22}$), indicating that the depth discrepancy between branches
reflects the presence of dynamic content across diverse scenes. Qualitative
examples in \cref{fig:raymap_compare} further show that the resulting
dynamic map closely aligns with ground-truth masks on both synthetic and real
scenes. We leverage this property to identify dynamic regions without
additional supervision.

\subsection{Dynamic Identification via RayMap Remap}
\label{ssec:remapping}

\begin{figure}[t]
    \centering
    \includegraphics[width=\linewidth]{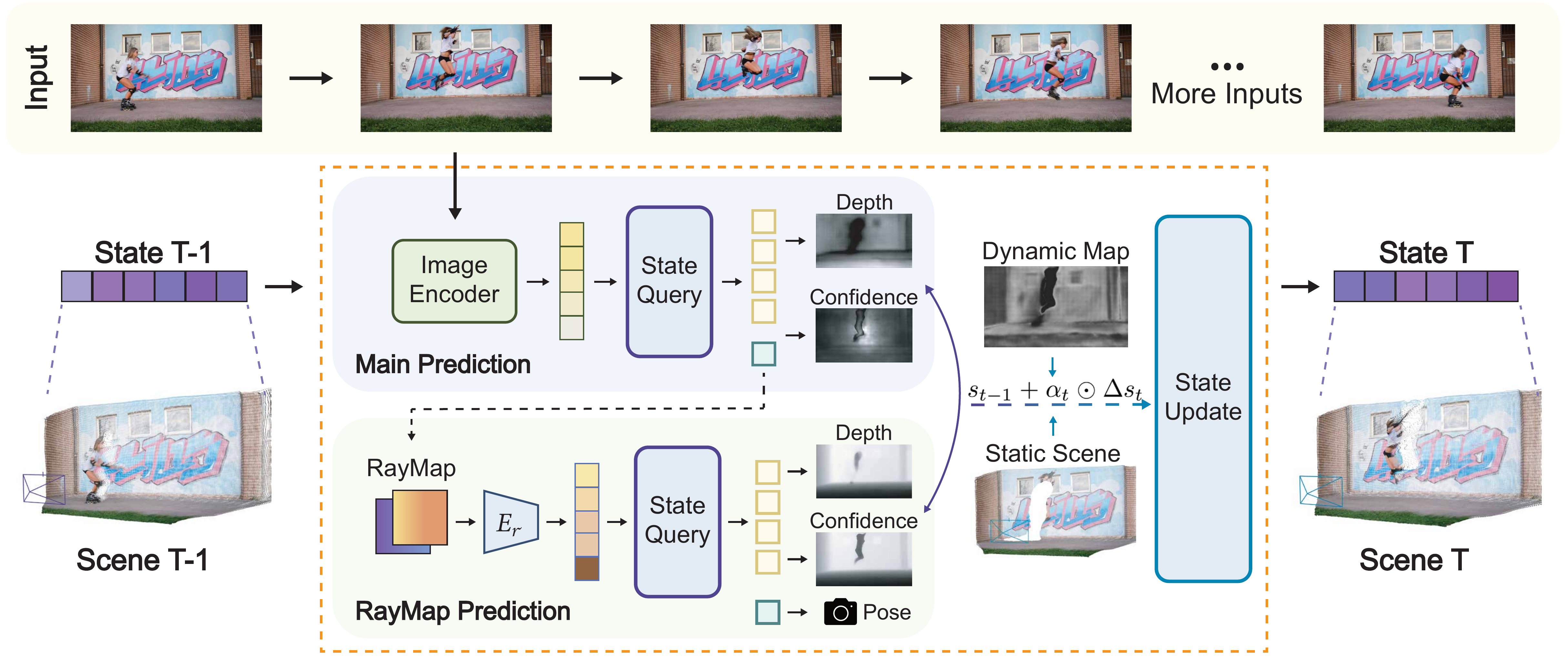}
    \caption{\textbf{Method Overview.}
    Our method performs streaming 3D reconstruction via dual-branch inference.
    At time step $t$, the \textbf{main branch} predicts depth and pose from state $s_{t-1}$ using both image and RayMap features.
    The predicted pose $\hat{\mathbf{T}}_t$ is remapped into RayMap and encoded into tokens $r'_t$, then queried against the \textit{frozen} state $s_{t-1}$ by the \textbf{RayMap branch} to obtain a static-biased prediction.
    The depth difference between branches is projected via image-state attention to form staticness weights $\alpha_t$ that suppress dynamic regions during state update ($s_t = s_{t-1} + \alpha_t \odot \Delta s_t$).
    This dual-branch scheme identifies and suppresses dynamic regions at inference time.
    }
    \label{fig:overview}
    \vspace{-1.5em}
\end{figure}

Building on the static bias observed in \cref{ssec:observation}, we
propose a dual-branch inference scheme to identify dynamic regions
(\cref{fig:overview}). At each timestep, both branches decode from
the same frozen state $s_{t-1}$: the main branch processes image and
RayMap features, while the RayMap branch processes only RayMap features
constructed from the main branch's predicted pose. The per-pixel depth
discrepancy between branches then serves as a signal for dynamic identification.

Specifically, given the main branch's predicted pose
$\hat{\mathbf{T}}_t$, we construct a new RayMap
$\mathcal{R}(\hat{\mathbf{T}}_t)$ and feed it through the encoder to
obtain RayMap tokens $r'_t$. The decodings for the main and RayMap
branches are given respectively by:
\begin{equation}
\hat{\mathbf{y}}^{\text{main}}_t = \text{Dec}(f_t + r_t, s_{t-1}), \quad
\hat{\mathbf{y}}^{\text{raymap}}_t = \text{Dec}(r'_t, s_{t-1}),
\end{equation}
where $f_t$ and $r_t$ are the original image and RayMap tokens.
Each prediction $\hat{\mathbf{y}}_t$ contains a depth map $z_t$ and
a confidence map.

We identify dynamic regions by comparing the depth and confidence
maps from both branches. For each pixel $i$, we compute the absolute relative depth
difference $\delta_i = |z^{\text{main}}_i - z^{\text{raymap}}_i| /
|z^{\text{main}}_i|$, where higher values indicate greater likelihood of dynamic content.
Since the staticness weights must operate on state tokens rather than
pixels, we aggregate $\delta_i$ in two steps. Pixel-level scores are
first pooled into image-token-level scores $\delta^{\text{tok}}_k$
via confidence-weighted averaging within each patch, where the weights
are the confidence scores predicted by the main branch. These are then
projected onto per-state-token scores $\delta^{\text{state}}_j$ as a
weighted average over $\delta^{\text{tok}}_k$ using the decoder's
cross-attention weights $A_{jk}$ between state token $j$ and image
token $k$.
We convert $\delta^{\text{state}}$ to staticness weights
$\alpha_t \in \mathbb{R}^N$, where $N$ is the number of state tokens:
\begin{equation}
\alpha_t = \sigma\left( \gamma \cdot
(\text{median}(\delta^{\text{state}}) - \delta^{\text{state}}) \,/\,
\text{IQR}(\delta^{\text{state}}) \right),
\end{equation}
where $\sigma$ is the sigmoid function, $\gamma$ controls the
sensitivity of dynamic-static separation,
and $\text{IQR}$ is the interquartile range
as a robust scale estimate.
Tokens with high $\alpha_t$ receive full state updates while those
with low $\alpha_t$ are suppressed, effectively filtering dynamic
regions from memory.

Finally, we apply these weights to modulate state updates.
We accumulate $\alpha_t$ over time via exponential moving average to
improve temporal stability, yielding the final weights that gate the
state update:
$s_t = s_{t-1} + \alpha_t \odot \Delta s_t$, where $\Delta s_t$
is the state update produced by the decoder and $\odot$ denotes
element-wise multiplication.
Additionally, we use the pixel-level staticness map to construct a
weighted global feature that biases pose retrieval toward static
regions during memory updates.
After a warmup using only the main branch, this dual-branch
scheme operates at each timestep without backpropagation, enabling
dynamic-aware streaming reconstruction with constant memory usage.

\subsection{Reset Metric Alignment}
\label{ssec:reset}

Streaming reconstruction models maintain a persistent memory to
accumulate scene understanding over time.
However, the memory mechanism suffers from forgetting over
extended sequences, as new observations gradually interfere with earlier
context.
To mitigate this, memory-based methods typically employ periodic
resets: the memory state is cleared and reinitialized using a repeated
frame to maintain temporal continuity.
We observe that this reset process introduces a notable problem:
metric misalignment across segments.

The reset mechanism alters the memory state, causing the model to
produce inconsistent camera parameters and geometry for the same
repeated frame before and after reset.
This discrepancy manifests as scale mismatch between segments, which
propagates as systematic drift throughout subsequent frames,
accumulating error that degrades reconstruction quality over long
sequences.

We address this by exploiting the property that the repeated frame
should yield consistent reconstructions before and after reset.
Since both observations capture the same physical scene, any
discrepancy between their reconstructions directly reflects the metric
misalignment introduced by the reset process, providing a measurable
signal for correction.

Specifically, we estimate a Sim(3) transformation that aligns the
point cloud reconstructions from the repeated frame before and after
the reset, as both observations capture the same physical scene.
This transformation captures both the scale mismatch and pose offset
between segments.
The estimated transformation is then applied to all subsequent frames
in the new segment, restoring metric alignment and reducing error
accumulation.

\subsection{State-Aware Smoothing}
\label{ssec:smoothing}

Streaming methods estimate camera poses sequentially, which may lead to noisy predictions that accumulate over time. While post-hoc optimization~\cite{teed2021droid, duisterhof2024mast3r, teed2023deep} can refine trajectories offline, it is incompatible with online processing. To this end, we introduce state-aware smoothing, which derives
a per-frame smoothing coefficient from trajectory acceleration
and internal state change magnitude to adaptively stabilize
pose predictions online.

To derive the per-frame confidence, we define the state change
signal $\mathrm{sc}_t$ as the mean $\ell_2$ norm of $\Delta s_t$
across all $N$ state tokens, computed prior to the gated update
in \cref{ssec:remapping}, so that it reflects the full proposed
change. We further define the trajectory acceleration $a_t =
\|\mathbf{d}_t - \mathbf{d}_{t-1}\|_2$, where $\mathbf{d}_t =
\boldsymbol{\tau}_t - \boldsymbol{\tau}_{t-1}$ is the inter-frame
camera displacement. The product $a_t \times \mathrm{sc}_t$
captures motion irregularity and model uncertainty, reducing
false positives from either signal alone: a high $a_t$ with low
$\mathrm{sc}_t$ may indicate steady fast motion rather than noise,
and vice versa.

To suppress unreliable predictions while preserving stable motion
estimates, we convert the product $a_t \times \mathrm{sc}_t$ into a smoothing coefficient $\beta_t \in [0,1]$ via an inverse mapping and exponentially filter the inter-frame displacements:
\begin{equation}
    \hat{\mathbf{d}}_t = \beta_t \, \mathbf{d}_t + (1 - \beta_t) \, \hat{\mathbf{d}}_{t-1}, \quad \beta_t = \frac{1}{1 + \lambda \, |a_t \times \mathrm{sc}_t|},
\end{equation}
where $\boldsymbol{\tau}_t \in \mathbb{R}^3$ is the translation component of the predicted pose $\hat{\mathbf{T}}_t$, $\lambda$ controls the sensitivity of the mapping, and $\hat{\mathbf{d}}_0 = \mathbf{0}$. When both $a_t$ and $\mathrm{sc}_t$ are large, $\beta_t$ approaches zero and the filter relies on the accumulated estimate $\hat{\mathbf{d}}_{t-1}$, attenuating trajectory jitter; when the product is small, $\beta_t$ approaches one and the raw displacement is preserved. The filtered position $\hat{\boldsymbol{\tau}}_t = \hat{\boldsymbol{\tau}}_{t-1} + \hat{\mathbf{d}}_t$ is computed recursively, producing a smoothed trajectory without explicit history storage. Expanding the recursion yields the closed-form trajectory:
\begin{equation}
    \hat{\boldsymbol{\tau}}_t = \boldsymbol{\tau}_0 + \sum_{m=1}^{t} \sum_{k=1}^{m} \beta_k \prod_{l=k+1}^{m} (1 - \beta_l) \cdot \mathbf{d}_k,
\end{equation}
where $\boldsymbol{\tau}_0$ is the initial translation. The product $\beta_k \prod_{l=k+1}^{m}(1-\beta_l)$ governs the effective contribution of each past displacement $\mathbf{d}_k$, forming an adaptive decay that concentrates weight on recent frames during stable intervals and broadens the filtering horizon during high-uncertainty intervals. The inverse mapping responds to the absolute magnitude of $a_t \times \mathrm{sc}_t$, so sustained fast motion at constant velocity produces low acceleration and leaves $\beta_t$ close to one, avoiding excessive smoothing. This causal scheme operates fully online with negligible overhead.

\section{Experiments}
\label{sec:experiments}

We evaluate RayMap3R on three core 3D tasks: video depth estimation (\cref{subsec:depth}), camera pose estimation (\cref{subsec:pose}), and 3D reconstruction (\cref{subsec:recon}).

\noindent\textbf{Baselines.}
We compare RayMap3R against streaming 3D reconstruction methods
including Spann3R~\cite{spann3r}, CUT3R~\cite{cut3r},
Point3R~\cite{point3r}, StreamVGGT~\cite{zhuo2025streaming}, and
TTT3R~\cite{ttt3r}.
Spann3R extends DUSt3R~\cite{wang2024dust3r} with spatial memory
mechanisms for efficient reconstruction.
CUT3R maintains an implicit memory cache for continuous
reconstruction.
Point3R enhances CUT3R with explicit point anchors.
StreamVGGT builds upon VGGT~\cite{wang2025vggt} through knowledge
distillation to enable streaming reconstruction.
TTT3R extends CUT3R by using cross-attention as soft-gate guidance
for long-range reconstruction.
We also evaluate against offline reconstruction
methods~\cite{wang2025vggt, wang2024dust3r, duisterhof2024mast3r, monst3r},
which achieve higher accuracy but are not viable for long sequences
due to prohibitive memory growth.

\noindent\textbf{Datasets.}
Following previous work~\cite{cut3r,monst3r, ttt3r}, we evaluate on
diverse benchmarks covering both dynamic and static scenes.
For dynamic scenes, we use Sintel~\cite{butler2012naturalistic},
TUM-Dynamics~\cite{sturm2012benchmark},
KITTI~\cite{geiger2013vision}, and Bonn~\cite{palazzolo2019refusion}.
For static scenes, we evaluate on ScanNet~\cite{dai2017scannet} and
7-Scenes~\cite{shotton2013scene}. These benchmarks collectively cover diverse conditions including
dynamic and static, synthetic and real-world scenes.

\begin{table}[t]
\caption{\textbf{Video Depth Estimation.} We report scale-invariant
relative depth (per-sequence scale alignment) and metric-scale
absolute depth accuracy. RayMap3R achieves leading depth
accuracy among streaming methods under both settings.
\xmark\ denotes offline methods; \cmark\ denotes streaming methods.}
\vspace{-0.5em}

\label{tab:video_depth}
\centering
\renewcommand{\arraystretch}{1.02}
\renewcommand{\tabcolsep}{1.5pt}
\resizebox{\linewidth}{!}{
\begin{tabular}{@{}ll@{\hskip 0pt}c
>{\centering\arraybackslash}p{1.4cm}>{\centering\arraybackslash}p{1.35cm}
|>{\centering\arraybackslash}p{1.4cm}>{\centering\arraybackslash}p{1.35cm}
|>{\centering\arraybackslash}p{1.4cm}>{\centering\arraybackslash}p{1.35cm}@{}}
\toprule
 &  &
 & \multicolumn{2}{c}{\textbf{KITTI~\cite{geiger2013vision}}}
 & \multicolumn{2}{c}{\textbf{BONN~\cite{palazzolo2019refusion}}}
 & \multicolumn{2}{c}{\textbf{Sintel~\cite{butler2012naturalistic}}} \\
\cmidrule(lr){4-5} \cmidrule(lr){6-7} \cmidrule(lr){8-9}
\textbf{Alignment} & \textbf{Method}
& \textbf{Onl.} & {Abs Rel $\downarrow$} & {$\delta$\textless$1.25\uparrow$}
& {Abs Rel $\downarrow$} & {$\delta$\textless$1.25\uparrow$}
& {Abs Rel $\downarrow$} & {$\delta$\textless$1.25\uparrow$} \\
\midrule

\multirow{11}{*}{\begin{minipage}{2cm}Per-sequence\\scale\end{minipage}}

& \textcolor{gray!60}{DUSt3R~\cite{wang2024dust3r}} & \textcolor{gray!60}{\xmark}
& \textcolor{gray!60}{0.144} & \textcolor{gray!60}{81.3}
& \textcolor{gray!60}{0.155} & \textcolor{gray!60}{83.3}
& \textcolor{gray!60}{0.656} & \textcolor{gray!60}{45.2} \\

& \textcolor{gray!60}{MonST3R~\cite{monst3r}} & \textcolor{gray!60}{\xmark}
& \textcolor{gray!60}{0.168} & \textcolor{gray!60}{74.4}
& \textcolor{gray!60}{0.067} & \textcolor{gray!60}{96.3}
& \textcolor{gray!60}{0.378} & \textcolor{gray!60}{55.8} \\

& \textcolor{gray!60}{VGGT~\cite{wang2025vggt}} & \textcolor{gray!60}{\xmark}
& \textcolor{gray!60}{\textbf{0.070}} & \textcolor{gray!60}{\textbf{96.5}}
& \textcolor{gray!60}{\textbf{0.055}} & \textcolor{gray!60}{\textbf{97.1}}
& \textcolor{gray!60}{\textbf{0.287}} & \textcolor{gray!60}{\textbf{66.1}} \\

\cmidrule{2-9}
& Spann3R~\cite{spann3r} & \cmark
& 0.198 & 73.7
& 0.144 & 81.3
& 0.622 & 42.6 \\
& Point3R~\cite{point3r} & \cmark
& 0.135 & 84.0
& \secondc{0.061} & 96.2
& 0.451 & 48.7 \\
& StreamVGGT~\cite{zhuo2025streaming} & \cmark
& 0.173 & 72.1
& 0.063 & \secondc{97.2}
& \firstc{0.323} & \firstc{65.7} \\
& CUT3R~\cite{cut3r} & \cmark
& 0.118 & 88.1
& 0.078 & 93.7
& 0.421 & 47.9 \\
& TTT3R~\cite{ttt3r} & \cmark
& \secondc{0.114} & \secondc{90.4}
& 0.068 & 95.4
& 0.409 & 48.8 \\
& \textbf{Ours} & \cmark
& \firstc{0.098} & \firstc{92.8}
& \firstc{0.057} & \firstc{97.4}
& \secondc{0.401} & \secondc{50.9} \\

\midrule
\multirow{4}{*}{\begin{minipage}{2cm}Metric scale\end{minipage}}
& Point3R~\cite{point3r} & \cmark
& 0.190 & 73.9
& 0.136 & \secondc{94.6}
& \firstc{0.778} & 17.0 \\
& CUT3R~\cite{cut3r} & \cmark
& 0.122 & 85.5
& 0.103 & 88.5
& 1.029 & \secondc{23.8} \\
& TTT3R~\cite{ttt3r} & \cmark
& \secondc{0.111} & \secondc{88.8}
& \secondc{0.089} & 94.2
& 0.977 & 23.2 \\
& \textbf{Ours} & \cmark
& \firstc{0.104} & \firstc{89.4}
& \firstc{0.085} & \firstc{94.8}
& \secondc{0.954} & \firstc{24.0} \\

\bottomrule
\end{tabular}
}
\vspace{-1em}
\end{table}

\subsection{Video Depth Estimation}
\label{subsec:depth}
Following previous works~\cite{monst3r, cut3r, ttt3r}, we evaluate
video depth estimation on Sintel~\cite{butler2012naturalistic},
KITTI~\cite{geiger2013vision}, and Bonn~\cite{palazzolo2019refusion},
covering dynamic and static scenes across indoor and outdoor
environments. We use absolute relative error (Abs Rel) and
$\delta < 1.25$ as metrics. Following~\cite{cut3r, ttt3r}, we
report results under two protocols: per-sequence scale alignment,
which evaluates relative depth accuracy, and metric scale without
alignment, which measures absolute scale consistency.

As shown in \cref{tab:video_depth}, RayMap3R achieves leading
performance among streaming methods under both evaluation protocols.
Under per-sequence scale alignment, our method leads on Bonn and KITTI, while on Sintel StreamVGGT achieves lower depth error than ours, albeit with memory that scales with sequence length, among the compared streaming baselines.
The improvement on KITTI is pronounced, as
explicitly querying static regions via RayMap yields more temporally
consistent depth predictions in large-scale outdoor scenes. Under
the metric-scale setting, RayMap3R leads on Bonn and KITTI, and
achieves competitive accuracy on Sintel, where Point3R obtains
lower absolute error but degrades on threshold accuracy. RayMap3R
maintains stable performance across both protocols, suggesting that
suppressing dynamic regions during state updates helps preserve
scale consistency otherwise corrupted by dynamic regions.

\subsection{Camera Pose Estimation}
\label{subsec:pose}
Following previous works~\cite{cut3r, chen2024leap, monst3r}, we
evaluate camera pose estimation on Sintel~\cite{butler2012naturalistic}
with complex dynamic content, TUM-dynamics~\cite{sturm2012benchmark}
with real-world dynamic scenes, and ScanNet~\cite{dai2017scannet}
for static indoor environments. We report Absolute Translation
Error~(ATE) for global trajectory accuracy and Relative Pose
Error~(RPE) for translation and rotation using Sim(3) alignment.

As shown in \cref{tab:camera_pose}, RayMap3R achieves leading ATE and translational RPE across all datasets, while maintaining strong rotational RPE. On Sintel, where dynamic objects occupy large
portions of the frame, our dual-branch scheme effectively identifies
and suppresses their interference, yielding
substantial ATE improvement over the next-best streaming method.
On TUM-dynamics, which contains real-world moving objects with
diverse motion patterns, our method similarly achieves leading
trajectory accuracy, suggesting that the static bias of RayMap
predictions generalizes across different dynamic scenarios. CUT3R
achieves comparable rotational RPE on ScanNet yet suffers from
accumulated trajectory drift reflected by higher ATE, which our
reset metric alignment and state-aware smoothing help to mitigate.
On static ScanNet, our method matches the leading streaming method
in ATE, suggesting that dynamic filtering via staticness weights
does not degrade performance in the absence of moving objects.

\begin{table}[t]
\caption{
\textbf{Camera Pose Estimation.} RayMap3R achieves leading
trajectory accuracy among streaming methods across all datasets,
with particularly strong performance on dynamic sequences, while
remaining competitive on static scenes.
}
\vspace{-0.5em}
\label{tab:camera_pose}
\centering
\footnotesize
\renewcommand{\arraystretch}{1.02}
\renewcommand{\tabcolsep}{1.5pt}
\resizebox{\linewidth}{!}{
\begin{tabular}{@{}l l@{\hskip 3pt}c@{\hskip 3pt} ccc|ccc|ccc@{}}
\toprule
 &    &
 & \multicolumn{3}{c}{\makebox[3.5cm][c]{\textbf{Sintel~\cite{butler2012naturalistic}}}}
 & \multicolumn{3}{c}{\makebox[3.5cm][c]{\textbf{TUM-dyn~\cite{sturm2012benchmark}}}}
 & \multicolumn{3}{c}{\makebox[3.5cm][c]{\textbf{ScanNet~\cite{dai2017scannet}}}} \\
\cmidrule(lr){4-6} \cmidrule(lr){7-9} \cmidrule(lr){10-12}
 & \textbf{Method} & \textbf{Onl.} 
 & {ATE $\downarrow$} & {RPE$_t$ $\downarrow$} & {RPE$_r$ $\downarrow$}
 & {ATE $\downarrow$} & {RPE$_t$ $\downarrow$} & {RPE$_r$ $\downarrow$}
 & {ATE $\downarrow$} & {RPE$_t$ $\downarrow$} & {RPE$_r$ $\downarrow$} \\
\midrule

 & \cg DUSt3R~\cite{wang2024dust3r} & \cg \xmark
 & \cg 0.417 & \cg 0.250 & \cg 5.796
 & \cg 0.083 & \cg 0.017 & \cg 3.567
 & \cg 0.081 & \cg 0.028 & \cg 0.784 \\

 & \cg MASt3R~\cite{duisterhof2024mast3r} & \cg \xmark
 & \cg 0.185 & \cg 0.060 & \cg 1.496
 & \cg {0.038} & \cg {0.012} & \cg {0.448}
 & \cg 0.078 & \cg 0.020 & \cg {0.475} \\

 & \cg MonST3R~\cite{monst3r} & \cg \xmark
 & \cg \bf {0.111} & \cg \bf 0.044 & \cg 0.869
 & \cg 0.098 & \cg 0.019 & \cg 0.935
 & \cg 0.077 & \cg 0.018 & \cg 0.529 \\

 & \cg VGGT~\cite{wang2025vggt} & \cg \xmark
 & \cg 0.172 & \cg 0.062 & \cg \textbf{0.471}
 & \cg \textbf{0.012} & \cg \textbf{0.010} & \cg \textbf{0.310}
 & \cg \textbf{0.035} & \cg \textbf{0.015} & \cg \textbf{0.377} \\

\midrule

& Spann3R~\cite{spann3r} & \cmark
& 0.329 & 0.110 & 4.471
& 0.056 & 0.021 & 0.591
& 0.096 & 0.023 & 0.661 \\

& Point3R~\cite{point3r} & \cmark
& 0.351 & 0.128 & 1.822
& 0.075 & 0.029 & 0.642
& 0.106 & 0.035 & 1.946 \\

& StreamVGGT~\cite{zhuo2025streaming} & \cmark
& 0.251 & 0.149 & 1.894
& 0.061 & 0.033 & 3.209
& 0.161 & 0.057 & 3.647 \\

& CUT3R~\cite{cut3r} & \cmark
& \secondc{0.208} & \secondc{0.072} & \firstc{0.636}
& {0.031} & 0.009 & 0.303
& 0.098 & {0.022} & \firstc{0.600} \\

& TTT3R~\cite{ttt3r} & \cmark
& 0.210 & {0.091} & 0.722
& \secondc{0.019} & \secondc{0.008} & \secondc{0.292}
& \secondc{0.065} & \secondc{0.021} & {0.637} \\

& \textbf{Ours} & \cmark
& \firstc{0.166} & \firstc{0.056} & \secondc{0.720}
& \firstc{0.018} & \firstc{0.005} & \firstc{0.287}
& \firstc{0.064} & \firstc{0.016} & \secondc{0.635} \\

\bottomrule
\end{tabular}
}
\end{table}
\begin{table}[t]
\caption{
\textbf{3D Reconstruction on 7-Scenes~\cite{shotton2013scene}.}
Each metric reports Mean, Median, and Minimum across scenes.
RayMap3R achieves leading reconstruction quality among streaming
methods across the majority of metrics.
}
\vspace{-0.5em}
\label{tab:3d_recon}
\centering
\footnotesize
\renewcommand{\arraystretch}{1.02}
\renewcommand{\tabcolsep}{1.25pt}
\resizebox{\linewidth}{!}{
\begin{tabular}{@{}l@{\hskip -3pt}c ccc|ccc|ccc|ccc@{}}
\toprule
&
& \multicolumn{3}{c}{{Acc}$\downarrow$}
& \multicolumn{3}{c}{{Comp}$\downarrow$}
& \multicolumn{3}{c}{{NC}$\uparrow$}
& \multicolumn{3}{c}{{Chamfer}$\downarrow$} \\
\cmidrule(lr){3-5} \cmidrule(lr){6-8} \cmidrule(lr){9-11} \cmidrule(lr){12-14}
\textbf{Method} & \textbf{Onl.}
& {Mean} & {Med.} & {Min}
& {Mean} & {Med.} & {Min}
& {Mean} & {Med.} & {Min}
& {Mean} & {Med.} & {Min} \\
\midrule
\textcolor{gray!60}{DUSt3R-GA~\cite{wang2024dust3r}} & \textcolor{gray!60}{\xmark}
& \textcolor{gray!60}{\bf 0.146} & \textcolor{gray!60}{\bf 0.077} & \textcolor{gray!60}{\bf 0.052}
& \textcolor{gray!60}{0.181}     & \textcolor{gray!60}{\bf 0.067} & \textcolor{gray!60}{\bf 0.042}
& \textcolor{gray!60}{\bf 0.736} & \textcolor{gray!60}{\bf 0.839} & \textcolor{gray!60}{\bf 0.865}
& \textcolor{gray!60}{\bf 0.327} & \textcolor{gray!60}{\bf 0.144} & \textcolor{gray!60}{\bf 0.094} \\

\textcolor{gray!60}{MASt3R-GA~\cite{duisterhof2024mast3r}} & \textcolor{gray!60}{\xmark}
& \textcolor{gray!60}{0.185}     & \textcolor{gray!60}{0.081}     & \textcolor{gray!60}{0.056}
& \textcolor{gray!60}{\bf 0.180} & \textcolor{gray!60}{0.069}     & \textcolor{gray!60}{0.047}
& \textcolor{gray!60}{0.701}     & \textcolor{gray!60}{0.792}     & \textcolor{gray!60}{0.821}
& \textcolor{gray!60}{0.365}     & \textcolor{gray!60}{0.150}     & \textcolor{gray!60}{0.103} \\

\textcolor{gray!60}{MonST3R-GA~\cite{monst3r}} & \textcolor{gray!60}{\xmark}
& \textcolor{gray!60}{0.248}     & \textcolor{gray!60}{0.185}     & \textcolor{gray!60}{0.142}
& \textcolor{gray!60}{0.266}     & \textcolor{gray!60}{0.167}     & \textcolor{gray!60}{0.121}
& \textcolor{gray!60}{0.672}     & \textcolor{gray!60}{0.759}     & \textcolor{gray!60}{0.783}
& \textcolor{gray!60}{0.514}     & \textcolor{gray!60}{0.352}     & \textcolor{gray!60}{0.263} \\
\midrule
Spann3R~\cite{spann3r} & \cmark
& 0.298 & 0.226 & 0.170
& 0.205 & 0.112 & 0.078
& \firstc{0.650} & \firstc{0.730} & \firstc{0.754}
& 0.503 & 0.338 & 0.248 \\

CUT3R~\cite{cut3r} & \cmark
& 0.043 & 0.026 & 0.015
& 0.031 & 0.018 & 0.009
& 0.621 & 0.618 & 0.523
& 0.027 & 0.025 & 0.013 \\

TTT3R~\cite{ttt3r} & \cmark
& \secondc{0.027}           & \secondc{0.024} & \secondc{0.012}
& \secondc{0.023} & \secondc{0.017}           & \firstc{0.005}
& 0.582           & 0.583           & 0.545
& \secondc{0.025} & \firstc{0.020}  & \secondc{0.011} \\

\textbf{Ours} & \cmark
& \firstc{0.023}  & \firstc{0.023}           & \firstc{0.009}
& \firstc{0.022}  & \firstc{0.016} & \secondc{0.007}
& \secondc{0.629} & \secondc{0.626} & \secondc{0.605}
& \firstc{0.024}  & \secondc{0.021} & \firstc{0.008} \\
\bottomrule
\end{tabular}
}
\vspace{-1em}
\end{table}

\subsection{3D Reconstruction}
\label{subsec:recon}
Following~\cite{cut3r, ttt3r}, we evaluate 3D reconstruction on
7-Scenes~\cite{shotton2013scene} using accuracy, completion, normal
consistency, and Chamfer distance, running 200 frames per scene. As
shown in \cref{tab:3d_recon}, RayMap3R achieves leading reconstruction
quality among streaming methods across the majority of metrics. In
particular, our method achieves the lowest accuracy error across mean,
median, and minimum statistics among all streaming approaches. Compared
to CUT3R, RayMap3R yields consistent improvements in both accuracy and
completion across all statistics, as filtering dynamic content before
state updates prevents corrupted geometry from accumulating in memory.
Against TTT3R, RayMap3R achieves lower accuracy across all statistics
and better completion on mean and median, with improved Chamfer on mean
and minimum and comparable median. Although Spann3R attains higher
normal consistency, RayMap3R outperforms it substantially on all other
geometric metrics, reflecting the benefit of dynamic-aware memory
updates. Furthermore, RayMap3R surpasses MonST3R-GA in accuracy and
Chamfer distance, suggesting that selective state updates can partially
compensate for the absence of global optimization in streaming
reconstruction.

\begin{figure}[t]
    \centering
    \includegraphics[width=\linewidth]{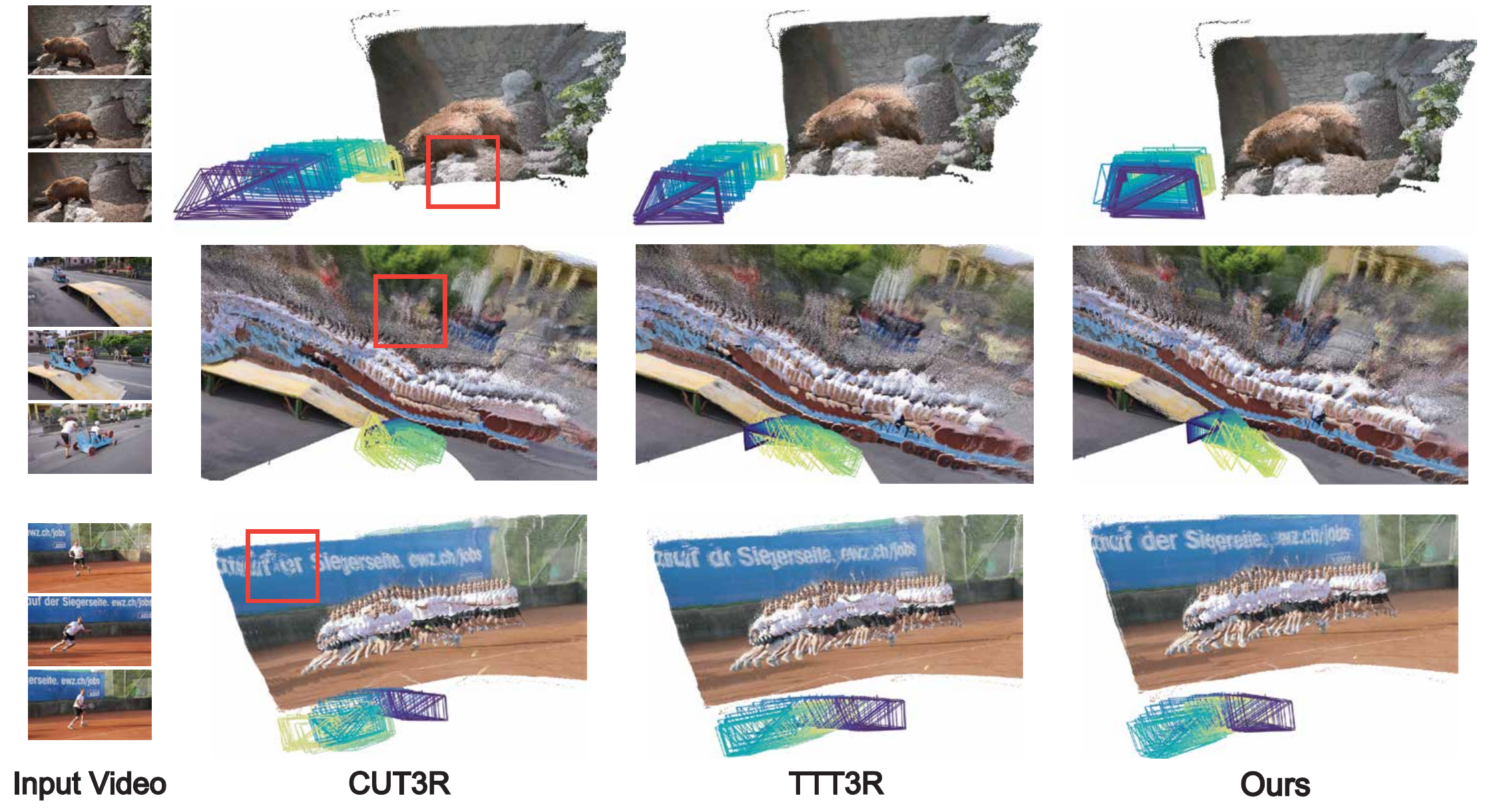}
    \caption{\textbf{Qualitative Results on DAVIS~\cite{perazzi2016benchmark} Videos.} We compare our method with CUT3R~\cite{cut3r} and TTT3R~\cite{ttt3r}. Our method achieves more stable camera pose estimation and produces clearer reconstructions.
    }
    \label{fig:qualitative}
    \vspace{-1em}
\end{figure}

\subsection{Qualitative Results}
We compare our method with CUT3R~\cite{cut3r} and TTT3R~\cite{ttt3r}
on dynamic sequences from the DAVIS dataset~\cite{perazzi2016benchmark}
as shown in \cref{fig:qualitative}. CUT3R suffers from noticeable
camera drift as dynamic content corrupts the memory state, leading
to distorted structures and misaligned surfaces. TTT3R improves
stability but still drifts when large dynamic regions dominate the
scene, as soft gating cannot fully suppress their influence on state
updates. In contrast, RayMap3R produces smoother trajectories and
more geometrically faithful reconstructions across all sequences,
demonstrating consistent robustness to diverse dynamic scenarios.
Notably, in the bottom row, the text on the boat surface is clearly
legible in our reconstruction, whereas CUT3R collapses the geometry
and TTT3R produces blurred results, illustrating that static-region
querying preserves fine-grained geometric details that dynamic
interference would otherwise corrupt.

\section{Analysis}

\textbf{Inference Speed and Memory Usage.}
We evaluate inference speed and GPU memory on ScanNet
using a single NVIDIA RTX A6000 48GB GPU.
As shown in \cref{tab:fps_memory}, offline
methods~\cite{monst3r, wang2025vggt} require all frames before
processing and are not viable for streaming scenarios;
their memory already exceeds 20\,GB at 50 views and grows
prohibitively beyond 200 views.
Point3R~\cite{point3r} is online but maintains an explicit spatial
pointer memory that grows with sequence length, resulting in
reduced throughput and out-of-memory beyond 900
views.
RayMap3R adopts the implicit-state design, maintaining stable memory across all sequence lengths; the dual-branch design introduces a moderate speed overhead, though accuracy gains are consistent across pose and depth benchmarks (\cref{tab:video_depth,tab:camera_pose}).

\begin{table*}[t]
\centering
\begin{minipage}[t]{0.48\linewidth}
\caption{\textbf{Inference speed and GPU memory.}
RayMap3R achieves competitive throughput and reasonable memory.
}
\label{tab:fps_memory}
\centering
\footnotesize
\renewcommand{\arraystretch}{1.05}
\renewcommand{\tabcolsep}{3pt}
\resizebox{\linewidth}{!}{%
\begin{tabular}{@{}l|cc|cc@{}}
\toprule
\textbf{Method}
& \multicolumn{2}{c|}{\textbf{50 views}}
& \multicolumn{2}{c}{\textbf{1000 views}} \\
\cmidrule(lr){2-3} \cmidrule(lr){4-5}
 & {Mem $\downarrow$} & {FPS $\uparrow$}
 & {Mem $\downarrow$} & {FPS $\uparrow$} \\
\midrule
\textcolor{gray}{MonST3R~\cite{monst3r}}   & \textcolor{gray}{32.0} & \textcolor{gray}{0.31} & \textcolor{gray}{OOM} & \textcolor{gray}{OOM} \\
\textcolor{gray}{VGGT~\cite{wang2025vggt}} & \textcolor{gray}{20.0} & \textcolor{gray}{21.0} & \textcolor{gray}{OOM} & \textcolor{gray}{OOM} \\
Point3R~\cite{point3r}  & 30.0 & 5.0  & OOM  & OOM  \\
CUT3R~\cite{cut3r}      & 6.4  & 19.7 & 6.5  & 19.7 \\
TTT3R~\cite{ttt3r}      & 7.6  & 19.6 & 7.7  & 19.6 \\
\textbf{Ours}           & 9.2  & 13.8 & 9.4  & 13.8 \\
\bottomrule
\end{tabular}%
}
\end{minipage}
\hfill
\begin{minipage}[t]{0.48\linewidth}
\caption{\textbf{Component ablation.}
R: dual-branch; M: metric alignment; S: state-aware smoothing.
}
\label{tab:ablation_raymap3r}
\centering
\footnotesize
\renewcommand{\arraystretch}{1.05}
\renewcommand{\tabcolsep}{3pt}
\resizebox{\linewidth}{!}{%
\begin{tabular}{@{}l|ccccc@{}}
\toprule
& {Base} & {R} & {R+M} & {R+S} & {\textbf{Full}} \\
\midrule
\multirow{2}{*}{ATE $\downarrow$}
& \multicolumn{5}{l}{\textcolor{gray}{\scriptsize Camera Pose}} \\
& 0.114 & 0.113 & 0.110 & 0.084 & \textbf{0.081} \\
\midrule
\multirow{2}{*}{AbsRel $\downarrow$}
& \multicolumn{5}{l}{\textcolor{gray}{\scriptsize Video Depth}} \\
& 0.208 & 0.186 & 0.184 & 0.185 & \textbf{0.183} \\
\midrule
\multirow{2}{*}{Chamfer $\downarrow$}
& \multicolumn{5}{l}{\textcolor{gray}{\scriptsize 3D Reconstruction}} \\
& 0.322    & 0.245    & 0.213    & 0.196    & \textbf{0.170}             \\
\bottomrule
\end{tabular}%
}
\end{minipage}
\vspace{-1em}
\end{table*}

\noindent\textbf{Component Ablation.}
We ablate three key components of RayMap3R using CUT3R as the
base: dual-branch RayMap identification~(R), reset metric
alignment~(M), and state-aware smoothing~(S), with results in
\cref{tab:ablation_raymap3r}. R yields the most substantial
improvement in depth and 3D reconstruction, as suppressing
dynamic interference reduces per-frame estimation error and
prevents corrupted geometry from accumulating in memory, with
modest effect on trajectory since filtering operates at the
state level. M produces notable Chamfer improvement by correcting
scale misalignment at reset boundaries, leading to more globally
consistent point clouds. S primarily improves trajectory accuracy
by adaptively filtering pose noise based on internal state
changes. The full model achieves consistent gains across all
tasks, indicating that R, M, and S address complementary aspects
of streaming reconstruction in dynamic scenes.



\noindent\textbf{Static Bias Analysis.}
We quantitatively evaluate the dynamic map across 108 sequences
using four metrics summarized in \cref{tab:static_bias}: disc
measures the mean discrepancy ratio between dynamic and static
regions; Mean AUC casts the signal as a binary classifier, where
0.5 denotes random chance; Spearman $\rho$ measures its
sequence-level correlation with ground-truth dynamic ratio; and
Mean IoU measures the overlap between the dynamic
map and ground-truth masks.

The discrepancy between main and RayMap-only predictions is
consistently larger in dynamic regions than static ones across datasets (disc $>$ 1), reflecting structured separation
rather than random variation. Mean AUC exceeds 0.5 on real-world
datasets, indicating reliable discriminative performance; the
lower AUC on Sintel is attributable to its more heterogeneous
motion patterns, which make threshold-based classification
harder. Spearman $\rho$ further shows that sequences with greater
dynamic content yield stronger detection, as illustrated in
\cref{fig:static_raymap} (right): despite its lower AUC, Sintel
achieves the highest $\rho$, reflecting that its wide
range of dynamic ratios produces a strong and consistent
ranking signal. The moderate IoU is expected, as the dynamic
map is a continuous signal thresholded against binary
ground-truth masks, designed as a soft gating cue rather than
a precise segmentation output. Together, these findings suggest
that the RayMap static bias generalizes reliably across diverse
scene conditions.
\begin{table}[t]
\centering
\vspace{-1.0em}
\caption{\textbf{Dynamic Map Evaluation.} Evaluation across 108
sequences from three datasets. The depth discrepancy signal shows
consistent discriminability and positive correlation with ground-truth
dynamics across diverse scenes.}
\label{tab:static_bias}
\scriptsize
\renewcommand{\arraystretch}{1.05}
\renewcommand{\tabcolsep}{3pt}
\resizebox{\linewidth}{!}{
\begin{tabular}{@{}lcc|cccc@{}}
\toprule
\textbf{Dataset} & \textbf{Seq} & \textbf{Frames} & \textbf{Mean disc} $\uparrow$ & \textbf{Mean AUC} $\uparrow$ & \textbf{Mean IoU} $\uparrow$ & $\rho$ $\uparrow$ \\
\midrule
MPI Sintel~\cite{butler2012naturalistic} & 19 & 746 & 1.61 & 0.464 & 0.298 & 0.900 \\
DAVIS 2017~\cite{perazzi2016benchmark} & 81 & 5205 & 1.68 & 0.538 & 0.189 & 0.713 \\
TUM RGB-D~\cite{sturm2012benchmark} & 8 & 680 & 1.88 & 0.560 & 0.206 & 0.643 \\
\midrule
All & 108 & 6631 & 1.69 & 0.532 & 0.203 & 0.771 \\
\bottomrule
\end{tabular}
}
\vspace{-1.0em}
\end{table}


\noindent\textbf{Discussion.}
RayMap3R relies on the model having developed implicit dynamic
awareness during training, which the static bias makes explicit.
This bias arises from the RayMap representation: encoding only
camera geometry without appearance cues forces the model to rely
on memory, which favors temporally consistent static structures
over transient dynamic objects. Nevertheless, if the training
distribution lacks dynamic scenes, the model may not develop
adequate sensitivity to distinguish the two, limiting the
effectiveness of the dual-branch scheme. Beyond the streaming
setting, as RayMap representations are increasingly adopted in
offline feed-forward models~\cite{da3, mapanything},
investigating analogous biases in offline architectures presents
a promising direction for training-free dynamic scene
understanding and reconstruction at broader scales.

\section{Conclusion}

We presented RayMap3R, a training-free framework for streaming 3D reconstruction in dynamic scenes. By exploiting the static bias in RayMap-only predictions, our dual-branch inference scheme identifies dynamic regions without additional supervision or retraining. Combined with reset metric alignment and state-aware smoothing, RayMap3R suppresses dynamic interference while preserving metric consistency and real-time efficiency. Extensive experiments demonstrate state-of-the-art performance among streaming methods across camera pose estimation, video depth prediction, and 3D reconstruction, with particularly strong improvements on dynamic sequences. We hope this work motivates further exploration of implicit geometric biases in feed-forward models as a resource for dynamic scene understanding and reconstruction.

\bibliographystyle{splncs04}
\bibliography{main}

@inproceedings{wang2024dust3r,
  title={Dust3r: Geometric 3d vision made easy},
  author={Wang, Shuzhe and Leroy, Vincent and Cabon, Yohann and Chidlovskii, Boris and Revaud, Jerome},
  booktitle={Proceedings of the IEEE/CVF Conference on Computer Vision and Pattern Recognition},
  pages={20697--20709},
  year={2024}
}

@article{lu2024align3r,
  title={Align3r: Aligned monocular depth estimation for dynamic videos},
  author={Lu, Jiahao and Huang, Tianyu and Li, Peng and Dou, Zhiyang and Lin, Cheng and Cui, Zhiming and Dong, Zhen and Yeung, Sai-Kit and Wang, Wenping and Liu, Yuan},
  journal={arXiv preprint arXiv:2412.03079},
  year={2024}
}

@article{li2024megasam,
  title={Megasam: Accurate, fast, and robust structure and motion from casual dynamic videos},
  author={Li, Zhengqi and Tucker, Richard and Cole, Forrester and Wang, Qianqian and Jin, Linyi and Ye, Vickie and Kanazawa, Angjoo and Holynski, Aleksander and Snavely, Noah},
  journal={arXiv preprint arXiv:2412.04463},
  year={2024}
}

@article{wang2025vggt,
  title={Vggt: Visual geometry grounded transformer},
  author={Wang, Jianyuan and Chen, Minghao and Karaev, Nikita and Vedaldi, Andrea and Rupprecht, Christian and Novotny, David},
  journal={arXiv preprint arXiv:2503.11651},
  year={2025}
}

@article{cut3r,
  title={Continuous 3D Perception Model with Persistent State},
  author={Wang, Qianqian and Zhang, Yifei and Holynski, Aleksander and Efros, Alexei A and Kanazawa, Angjoo},
  journal={arXiv preprint arXiv:2501.12387},
  year={2025}
}

@article{spann3r,
  title={3d reconstruction with spatial memory},
  author={Wang, Hengyi and Agapito, Lourdes},
  journal={arXiv preprint arXiv:2408.16061},
  year={2024}
}

@article{sucar2025dynamic,
  title={Dynamic Point Maps: A Versatile Representation for Dynamic 3D Reconstruction},
  author={Sucar, Edgar and Lai, Zihang and Insafutdinov, Eldar and Vedaldi, Andrea},
  journal={arXiv preprint arXiv:2503.16318},
  year={2025}
}

@inproceedings{palazzolo2019refusion,
  title={ReFusion: 3D reconstruction in dynamic environments for RGB-D cameras exploiting residuals},
  author={Palazzolo, Emanuele and Behley, Jens and Lottes, Philipp and Giguere, Philippe and Stachniss, Cyrill},
  booktitle={2019 IEEE/RSJ International Conference on Intelligent Robots and Systems (IROS)},
  pages={7855--7862},
  year={2019},
  organization={IEEE}
}

@inproceedings{butler2012naturalistic,
  title={A naturalistic open source movie for optical flow evaluation},
  author={Butler, Daniel J and Wulff, Jonas and Stanley, Garrett B and Black, Michael J},
  booktitle={Computer Vision--ECCV 2012: 12th European Conference on Computer Vision, Florence, Italy, October 7-13, 2012, Proceedings, Part VI 12},
  pages={611--625},
  year={2012},
  organization={Springer}
}

@article{geiger2013vision,
  title={Vision meets robotics: The kitti dataset},
  author={Geiger, Andreas and Lenz, Philip and Stiller, Christoph and Urtasun, Raquel},
  journal={The international journal of robotics research},
  volume={32},
  number={11},
  pages={1231--1237},
  year={2013},
  publisher={Sage Publications Sage UK: London, England}
}

@article{teed2023deep,
  title={Deep patch visual odometry},
  author={Teed, Zachary and Lipson, Lahav and Deng, Jia},
  journal={Advances in Neural Information Processing Systems},
  volume={36},
  pages={39033--39051},
  year={2023}
}

@article{mapanything,
  title={Mapanything: Universal feed-forward metric 3d reconstruction},
  author={Keetha, Nikhil and M{\"u}ller, Norman and Sch{\"o}nberger, Johannes and Porzi, Lorenzo and Zhang, Yuchen and Fischer, Tobias and Knapitsch, Arno and Zauss, Duncan and Weber, Ethan and Antunes, Nelson and others},
  journal={arXiv preprint arXiv:2509.13414},
  year={2025}
}

@article{da3,
  title={Depth anything 3: Recovering the visual space from any views},
  author={Lin, Haotong and Chen, Sili and Liew, Junhao and Chen, Donny Y and Li, Zhenyu and Shi, Guang and Feng, Jiashi and Kang, Bingyi},
  journal={arXiv preprint arXiv:2511.10647},
  year={2025}
}

@article{teed2021droid,
  title={Droid-slam: Deep visual slam for monocular, stereo, and rgb-d cameras},
  author={Teed, Zachary and Deng, Jia},
  journal={Advances in neural information processing systems},
  volume={34},
  pages={16558--16569},
  year={2021}
}

@article{duisterhof2024mast3r,
  title={MASt3R-SfM: a Fully-Integrated Solution for Unconstrained Structure-from-Motion},
  author={Duisterhof, Bardienus and Zust, Lojze and Weinzaepfel, Philippe and Leroy, Vincent and Cabon, Yohann and Revaud, Jerome},
  journal={arXiv preprint arXiv:2409.19152},
  year={2024}
}

@article{wimbauer2025anycam,
  title={AnyCam: Learning to Recover Camera Poses and Intrinsics from Casual Videos},
  author={Wimbauer, Felix and Chen, Weirong and Muhle, Dominik and Rupprecht, Christian and Cremers, Daniel},
  journal={arXiv preprint arXiv:2503.23282},
  year={2025}
}

@article{chen2025back,
  title={Back on Track: Bundle Adjustment for Dynamic Scene Reconstruction},
  author={Chen, Weirong and Zhang, Ganlin and Wimbauer, Felix and Wang, Rui and Araslanov, Nikita and Vedaldi, Andrea and Cremers, Daniel},
  journal={arXiv preprint arXiv:2504.14516},
  year={2025}
}

@article{agarwal2011building,
  title={Building rome in a day},
  author={Agarwal, Sameer and Furukawa, Yasutaka and Snavely, Noah and Simon, Ian and Curless, Brian and Seitz, Steven M and Szeliski, Richard},
  journal={Communications of the ACM},
  volume={54},
  number={10},
  pages={105--112},
  year={2011},
  publisher={ACM New York, NY, USA}
}

@inproceedings{li2018megadepth,
  title={Megadepth: Learning single-view depth prediction from internet photos},
  author={Li, Zhengqi and Snavely, Noah},
  booktitle={Proceedings of the IEEE conference on computer vision and pattern recognition},
  pages={2041--2050},
  year={2018}
}

@article{team2025aether,
  title={Aether: Geometric-aware unified world modeling},
  author={Team, Aether and Zhu, Haoyi and Wang, Yifan and Zhou, Jianjun and Chang, Wenzheng and Zhou, Yang and Li, Zizun and Chen, Junyi and Shen, Chunhua and Pang, Jiangmiao and others},
  journal={arXiv preprint arXiv:2503.18945},
  year={2025}
}

@article{ravi2024sam,
  title={Sam 2: Segment anything in images and videos},
  author={Ravi, Nikhila and Gabeur, Valentin and Hu, Yuan-Ting and Hu, Ronghang and Ryali, Chaitanya and Ma, Tengyu and Khedr, Haitham and R{\"a}dle, Roman and Rolland, Chloe and Gustafson, Laura and others},
  journal={arXiv preprint arXiv:2408.00714},
  year={2024}
}

@inproceedings{fast3r,
  title={Fast3r: Towards 3d reconstruction of 1000+ images in one forward pass},
  author={Yang, Jianing and Sax, Alexander and Liang, Kevin J and Henaff, Mikael and Tang, Hao and Cao, Ang and Chai, Joyce and Meier, Franziska and Feiszli, Matt},
  booktitle={Proceedings of the Computer Vision and Pattern Recognition Conference},
  pages={21924--21935},
  year={2025}
}

@article{vggtslam,
  title={Vggt-slam: Dense rgb slam optimized on the sl (4) manifold},
  author={Maggio, Dominic and Lim, Hyungtae and Carlone, Luca},
  journal={arXiv preprint arXiv:2505.12549},
  year={2025}
}

@article{ttt3r,
  title={Ttt3r: 3d reconstruction as test-time training},
  author={Chen, Xingyu and Chen, Yue and Xiu, Yuliang and Geiger, Andreas and Chen, Anpei},
  journal={arXiv preprint arXiv:2509.26645},
  year={2025}
}

@article{point3r,
  title={Point3R: Streaming 3D Reconstruction with Explicit Spatial Pointer Memory},
  author={Wu, Yuqi and Zheng, Wenzhao and Zhou, Jie and Lu, Jiwen},
  journal={arXiv preprint arXiv:2507.02863},
  year={2025}
}

@article{zhuo2025streaming,
  title={Streaming 4d visual geometry transformer},
  author={Zhuo, Dong and Zheng, Wenzhao and Guo, Jiahe and Wu, Yuqi and Zhou, Jie and Lu, Jiwen},
  journal={arXiv preprint arXiv:2507.11539},
  year={2025}
}

@article{vggtlong,
  title={VGGT-Long: Chunk it, Loop it, Align it--Pushing VGGT's Limits on Kilometer-scale Long RGB Sequences},
  author={Deng, Kai and Ti, Zexin and Xu, Jiawei and Yang, Jian and Xie, Jin},
  journal={arXiv preprint arXiv:2507.16443},
  year={2025}
}

@article{mur2015orb,
  title={ORB-SLAM: A versatile and accurate monocular SLAM system},
  author={Mur-Artal, Raul and Montiel, Jose Maria Martinez and Tardos, Juan D},
  journal={IEEE transactions on robotics},
  volume={31},
  number={5},
  pages={1147--1163},
  year={2015},
  publisher={IEEE}
}

@article{pollefeys2008detailed,
  title={Detailed real-time urban 3d reconstruction from video},
  author={Pollefeys, Marc and Nist{\'e}r, David and Frahm, J-M and Akbarzadeh, Amir and Mordohai, Philippos and Clipp, Brian and Engels, Chris and Gallup, David and Kim, S-J and Merrell, Paul and others},
  journal={International Journal of Computer Vision},
  volume={78},
  pages={143--167},
  year={2008},
  publisher={Springer}
}

@inproceedings{schonberger2016structure,
  title={Structure-from-motion revisited},
  author={Schonberger, Johannes L and Frahm, Jan-Michael},
  booktitle={Proceedings of the IEEE conference on computer vision and pattern recognition},
  pages={4104--4113},
  year={2016}
}

@article{davison2007monoslam,
  title={MonoSLAM: Real-time single camera SLAM},
  author={Davison, Andrew J and Reid, Ian D and Molton, Nicholas D and Stasse, Olivier},
  journal={IEEE transactions on pattern analysis and machine intelligence},
  volume={29},
  number={6},
  pages={1052--1067},
  year={2007},
  publisher={IEEE}
}

@article{campos2021orb,
  title={Orb-slam3: An accurate open-source library for visual, visual--inertial, and multimap slam},
  author={Campos, Carlos and Elvira, Richard and Rodr{\'\i}guez, Juan J G{\'o}mez and Montiel, Jos{\'e} MM and Tard{\'o}s, Juan D},
  journal={IEEE transactions on robotics},
  volume={37},
  number={6},
  pages={1874--1890},
  year={2021},
  publisher={IEEE}
}

@article{monst3r,
  title={Monst3r: A simple approach for estimating geometry in the presence of motion},
  author={Zhang, Junyi and Herrmann, Charles and Hur, Junhwa and Jampani, Varun and Darrell, Trevor and Cole, Forrester and Sun, Deqing and Yang, Ming-Hsuan},
  journal={arXiv preprint arXiv:2410.03825},
  year={2024}
}

@inproceedings{sturm2012benchmark,
  title={A benchmark for the evaluation of RGB-D SLAM systems},
  author={Sturm, J{\"u}rgen and Engelhard, Nikolas and Endres, Felix and Burgard, Wolfram and Cremers, Daniel},
  booktitle={2012 IEEE/RSJ international conference on intelligent robots and systems},
  pages={573--580},
  year={2012},
  organization={IEEE}
}

@inproceedings{dai2017scannet,
  title={Scannet: Richly-annotated 3d reconstructions of indoor scenes},
  author={Dai, Angela and Chang, Angel X and Savva, Manolis and Halber, Maciej and Funkhouser, Thomas and Nie{\ss}ner, Matthias},
  booktitle={Proceedings of the IEEE conference on computer vision and pattern recognition},
  pages={5828--5839},
  year={2017}
}

@inproceedings{chen2024leap,
  title={Leap-vo: Long-term effective any point tracking for visual odometry},
  author={Chen, Weirong and Chen, Le and Wang, Rui and Pollefeys, Marc},
  booktitle={Proceedings of the IEEE/CVF Conference on Computer Vision and Pattern Recognition},
  pages={19844--19853},
  year={2024}
}

@article{zhang2024cameras,
  title={Cameras as rays: Pose estimation via ray diffusion},
  author={Zhang, Jason Y and Lin, Amy and Kumar, Moneish and Yang, Tzu-Hsuan and Ramanan, Deva and Tulsiani, Shubham},
  journal={arXiv preprint arXiv:2402.14817},
  year={2024}
}

@inproceedings{ke2024repurposing,
  title={Repurposing diffusion-based image generators for monocular depth estimation},
  author={Ke, Bingxin and Obukhov, Anton and Huang, Shengyu and Metzger, Nando and Daudt, Rodrigo Caye and Schindler, Konrad},
  booktitle={Proceedings of the IEEE/CVF conference on computer vision and pattern recognition},
  pages={9492--9502},
  year={2024}
}

@inproceedings{ling2024dl3dv,
  title={Dl3dv-10k: A large-scale scene dataset for deep learning-based 3d vision},
  author={Ling, Lu and Sheng, Yichen and Tu, Zhi and Zhao, Wentian and Xin, Cheng and Wan, Kun and Yu, Lantao and Guo, Qianyu and Yu, Zixun and Lu, Yawen and others},
  booktitle={Proceedings of the IEEE/CVF Conference on Computer Vision and Pattern Recognition},
  pages={22160--22169},
  year={2024}
}

@inproceedings{sun2020scalability,
  title={Scalability in perception for autonomous driving: Waymo open dataset},
  author={Sun, Pei and Kretzschmar, Henrik and Dotiwalla, Xerxes and Chouard, Aurelien and Patnaik, Vijaysai and Tsui, Paul and Guo, James and Zhou, Yin and Chai, Yuning and Caine, Benjamin and others},
  booktitle={Proceedings of the IEEE/CVF conference on computer vision and pattern recognition},
  pages={2446--2454},
  year={2020}
}

@inproceedings{wangtartanair,
  title={Tartanair: A dataset to push the limits of visual slam. In 2020 IEEE},
  author={Wang, Wenshan and Zhu, Delong and Wang, Xiangwei and Hu, Yaoyu and Qiu, Yuheng and Wang, Chen and Hu, Yafei and Kapoor, Ashish and Scherer, Sebastian},
  booktitle={RSJ International Conference on Intelligent Robots and Systems (IROS)},
  pages={4909--4916},
  year={2020}
}

@inproceedings{yeshwanth2023scannet++,
  title={Scannet++: A high-fidelity dataset of 3d indoor scenes},
  author={Yeshwanth, Chandan and Liu, Yueh-Cheng and Nie{\ss}ner, Matthias and Dai, Angela},
  booktitle={Proceedings of the IEEE/CVF International Conference on Computer Vision},
  pages={12--22},
  year={2023}
}

@article{dehghan2021arkitscenes,
  title={ARKitScenes: A Diverse Real-World Dataset For 3D Indoor Scene Understanding Using Mobile RGB-D Data.},
  author={Dehghan, Afshin and Baruch, Gilad and Chen, Zhuoyuan and Feigin, Yuri and Fu, Peter and Gebauer, Thomas and Kurz, Daniel and Dimry, Tal and Joffe, Brandon and Schwartz, Arik and others},
  journal={NeurIPS Datasets and Benchmarks},
  volume={2},
  number={6},
  pages={16},
  year={2021}
}

@inproceedings{reizenstein2021common,
  title={Common objects in 3d: Large-scale learning and evaluation of real-life 3d category reconstruction},
  author={Reizenstein, Jeremy and Shapovalov, Roman and Henzler, Philipp and Sbordone, Luca and Labatut, Patrick and Novotny, David},
  booktitle={Proceedings of the IEEE/CVF international conference on computer vision},
  pages={10901--10911},
  year={2021}
}

@article{gao2024cat3d,
  title={Cat3d: Create anything in 3d with multi-view diffusion models},
  author={Gao, Ruiqi and Holynski, Aleksander and Henzler, Philipp and Brussee, Arthur and Martin-Brualla, Ricardo and Srinivasan, Pratul and Barron, Jonathan T and Poole, Ben},
  journal={arXiv preprint arXiv:2405.10314},
  year={2024}
}

@inproceedings{shotton2013scene,
  title={Scene coordinate regression forests for camera relocalization in RGB-D images},
  author={Shotton, Jamie and Glocker, Ben and Zach, Christopher and Izadi, Shahram and Criminisi, Antonio and Fitzgibbon, Andrew},
  booktitle={Proceedings of the IEEE conference on computer vision and pattern recognition},
  pages={2930--2937},
  year={2013}
}

@inproceedings{perazzi2016benchmark,
  title={A benchmark dataset and evaluation methodology for video object segmentation},
  author={Perazzi, Federico and Pont-Tuset, Jordi and McWilliams, Brian and Van Gool, Luc and Gross, Markus and Sorkine-Hornung, Alexander},
  booktitle={Proceedings of the IEEE conference on computer vision and pattern recognition},
  pages={724--732},
  year={2016}
}

\clearpage
\newpage
\appendix
\noindent\textbf{\Large Supplementary Material}
\vspace{0.5em}

The appendix provides implementation details and model backbone descriptions 
(Secs.~A--B), dynamic map visualizations and temporal consistency analysis 
(Secs.~C--D), quantitative analysis of the depth discrepancy signal across 
scene conditions (Sec.~E), a detailed description of reset metric alignment 
(Sec.~F) and ablation analysis of state-aware smoothing (Sec.~G), 
cross-dataset component ablation (Sec.~H), and a statement on LLM usage (Sec.~I).

\section*{A. Implementation Details}
Following previous works~\cite{cut3r, ttt3r}, we use official
implementations with default hyperparameters for all baseline methods
to facilitate fair comparison. All experiments are conducted on a
single NVIDIA RTX A6000 GPU with 48GB memory. RayMap3R builds upon the
pre-trained CUT3R backbone~\cite{cut3r} without any fine-tuning or
additional supervision, preserving the training-free nature of our
approach. For streaming reconstruction, the memory state is
periodically reset every 50 frames to mitigate state drift over long
sequences, with reset metric alignment applied at each reset boundary
as described in Sec.~F. For camera pose estimation, predicted
trajectories are aligned with ground truth via similarity
transformation. For depth evaluation, median scaling is applied per
sequence to remove scale ambiguity under the per-sequence alignment
protocol.

\section*{B. Model Backbone}
CUT3R~\cite{cut3r} is trained via a multi-stage curriculum on
32 diverse datasets spanning synthetic and real-world scenarios,
covering static and dynamic scenes, object-centric views, and
indoor/outdoor environments, including
CO3Dv2~\cite{reizenstein2021common},
ARKitScenes~\cite{dehghan2021arkitscenes},
ScanNet++~\cite{yeshwanth2023scannet++},
TartanAir~\cite{wangtartanair},
Waymo~\cite{sun2020scalability},
MegaDepth~\cite{li2018megadepth},
DL3DV~\cite{ling2024dl3dv},
and DynamicStereo~\cite{ke2024repurposing}.
Stage 1 establishes foundational geometry on static data at lower resolution; 
Stage 2 incorporates dynamic scenes with moving objects 
and partial annotations. As a result, the backbone develops implicit awareness of 
static-dynamic distinctions through the geometric-only nature of RayMap tokens. 
RayMap3R makes this implicit awareness explicit and actionable, exploiting it 
through training-free dual-branch inference to achieve dynamic-aware streaming 
reconstruction without extra supervision.

\section*{C. Dynamic Map Visualization}
\begin{figure}[htpb]
    \centering
    \includegraphics[width=\linewidth]{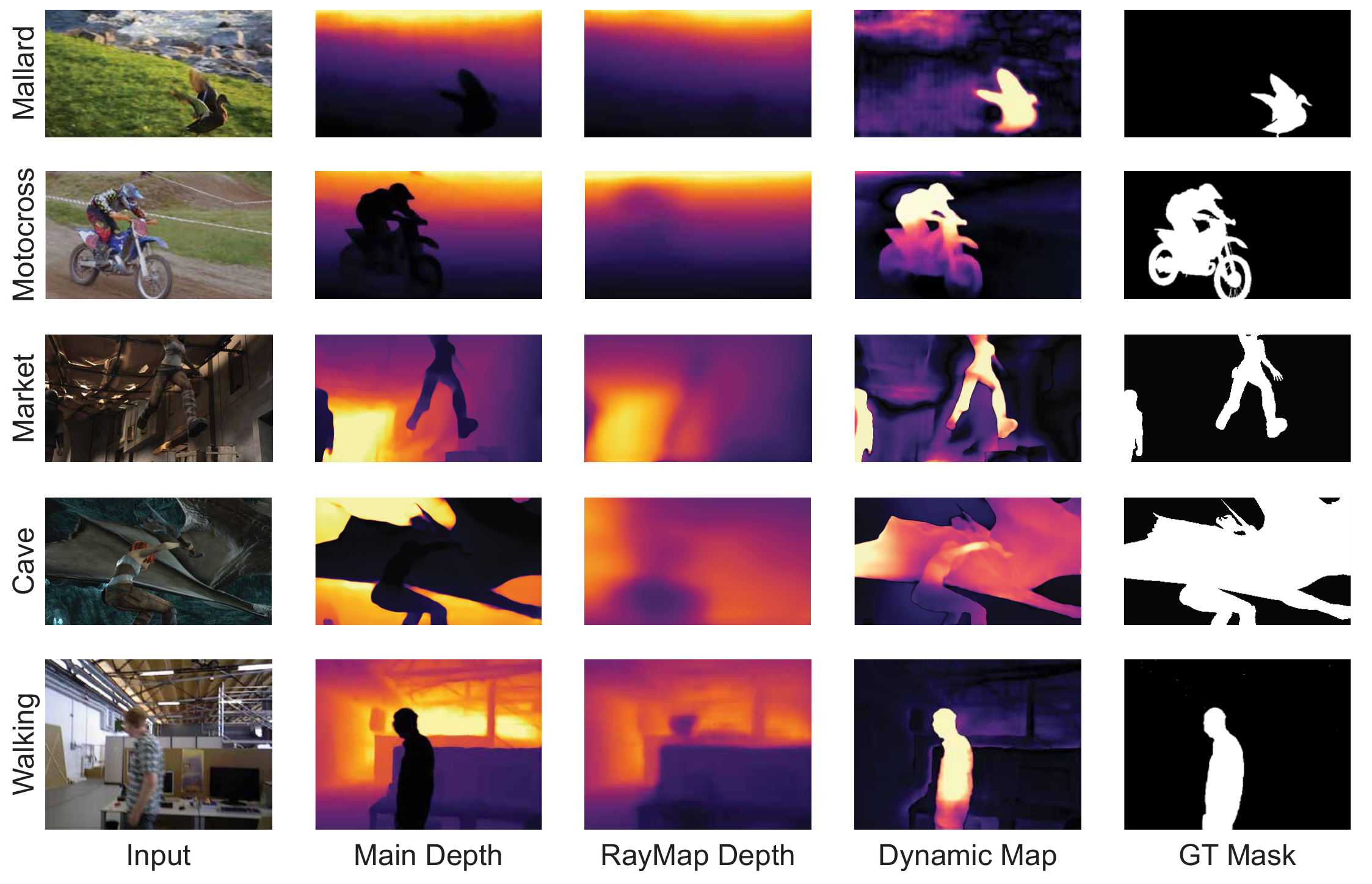}
\caption{
\textbf{Dynamic Map Visualization.}
We visualize the dynamic map inferred by RayMap3R across DAVIS (rows 1--2),
MPI Sintel (rows 3--4), and TUM Dynamic (row 5). The RayMap-only branch
suppresses dynamic objects relative to the main branch, and the resulting
dynamic map aligns with ground-truth masks.
}
    \label{fig:ablation_compare}

    \vspace{-1.0em}
\end{figure}

\cref{fig:ablation_compare} visualizes the dynamic map across
DAVIS~\cite{perazzi2016benchmark} (rows 1--2), MPI
Sintel~\cite{butler2012naturalistic} (rows 3--4), and TUM
Dynamic~\cite{sturm2012benchmark} (row 5), covering diverse real-world
and synthetic scenes with varied dynamic content, object scales, and
motion patterns.

The main branch depth incorporates both image and RayMap features and
captures full scene geometry including dynamic objects. The RayMap-only
depth exhibits a static bias: background structures such as terrain,
walls, and floors are reconstructed with reasonable fidelity and
depth boundaries, while dynamic objects tend to appear blurred or
geometrically inaccurate. This behavior arises because the RayMap
branch relies solely on geometric consistency encoded in memory rather
than per-frame appearance cues, causing it to favor temporally stable
structure over transient dynamic content.

The dynamic map, computed as the per-pixel depth discrepancy between
the two branches, highlights regions where the two predictions diverge.
On DAVIS, it responds to a small waterbird in motion against a
cluttered natural background and to a fast-moving motorcycle with
rider, suggesting sensitivity across a range of object scales. On
Sintel, moving figures and large foreground objects in cluttered
synthetic environments produce clear responses in the dynamic map,
while the surrounding static geometry remains relatively suppressed.
On TUM, a walking person is highlighted against a static indoor
background, with the dynamic map response concentrated on the person's
silhouette.

Across all rows, the dynamic map shows reasonable spatial
correspondence with the ground-truth masks without any mask
supervision, suggesting that the depth discrepancy between branches
provides a consistent proxy for dynamic content across real-world and
synthetic conditions.

\section*{D. Temporal Consistency of Dynamic Maps}
\begin{figure}[htpb]
    \centering
    \includegraphics[width=\linewidth]{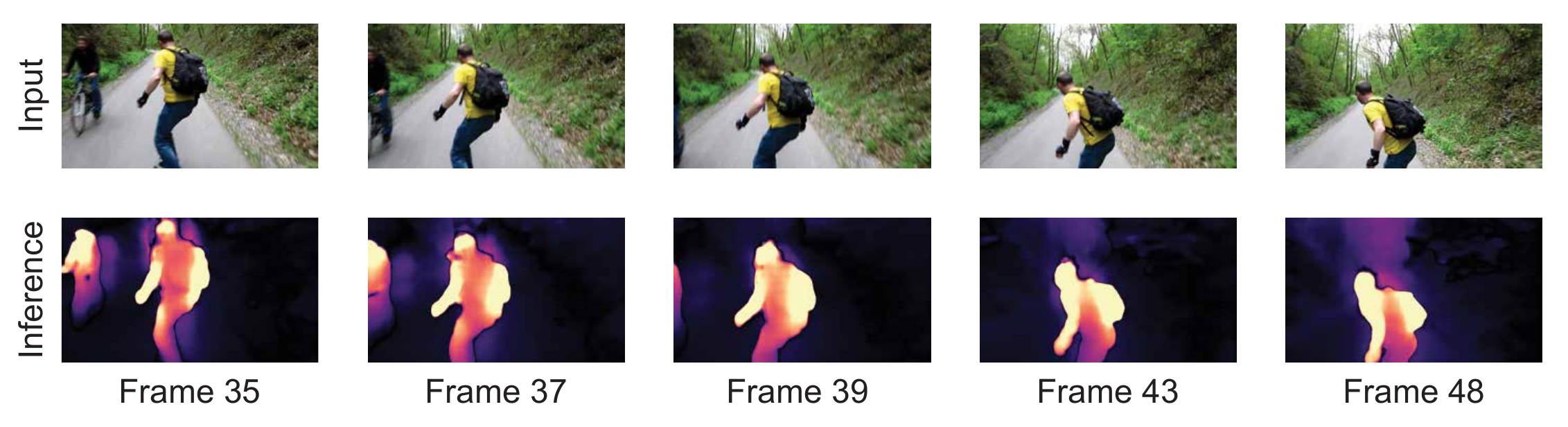}
\caption{
\textbf{Temporal Consistency of Dynamic Maps on DAVIS \textit{longboard}.}
Top: input frames. Bottom: dynamic maps.
The dynamic map consistently produces elevated responses at moving
subjects, including partially visible persons at the frame boundary.
}
    \label{fig:frames}

    \vspace{-2.0em}
\end{figure}

\cref{fig:frames} shows the dynamic map across five frames from the
DAVIS \textit{longboard} sequence, where multiple subjects move
continuously through the scene at varying distances from the camera.
Across all frames, the dynamic map consistently produces elevated
responses at moving subjects while assigning lower values to static
background regions such as the path surface and surrounding vegetation.

In frames 35 and 37, a partially visible person appears at the left
edge of the frame, and the dynamic map produces a clear response at
that location despite the limited spatial extent of the subject. The
response region follows each subject's spatial extent across frames
as position and apparent scale change. As the primary subject moves
further from the camera between frames 43 and 48, the response area
contracts correspondingly. While some background activations are
visible in certain frames, the signal remains predominantly
concentrated on the dynamic subjects throughout the sequence. These
observations suggest that the depth discrepancy between the main and
RayMap-only branches provides a temporally stable proxy for dynamic
content.

\section*{E. Dynamic Map Analysis across Scene Conditions}
To evaluate whether the depth discrepancy signal remains reliable
across varying levels of dynamic content, we stratify 131 sequences
from MPI Sintel~\cite{butler2012naturalistic}, DAVIS
2017~\cite{perazzi2016benchmark}, TUM
RGB-D~\cite{sturm2012benchmark}, and ScanNet~\cite{dai2017scannet} by
their mean ground-truth dynamic ratio into three groups: Low
($\leq$10\%, 66 sequences), Medium (10--30\%, 45 sequences), and High
($>$30\%, 20 sequences). This stratification spans the full spectrum
from predominantly static scenes to scenes with large dynamic
foreground objects, allowing us to assess signal behavior under diverse
conditions.

The mean $\delta$ increases monotonically across groups, consistent
with the expectation that overall depth discrepancy grows with dynamic
content. The disc metric exceeds 1 across all three groups, indicating
that dynamic pixels tend to produce higher discrepancy than static
pixels regardless of scene composition. The disc value in the Low
group is particularly notable: even when dynamic objects occupy less
than 10\% of the frame, the discrepancy in dynamic regions is
substantially larger than in static ones, suggesting that the separation is not solely driven by scene statistics. The decline in disc at higher dynamic
ratios is consistent with the observation that sustained exposure to
dynamic content can gradually corrupt the memory representation of
static regions, raising their baseline discrepancy and compressing the
ratio, while the absolute separation remains positive throughout. The
AUC exceeds 0.5 across all groups, suggesting consistent pixel-level
discriminability, with a modest decline at higher dynamic ratios that
may reflect increased scene complexity. The IoU increases monotonically
with dynamic ratio, which reflect the geometric
advantage of larger dynamic regions for threshold-based localization.

Together, these results suggest that the depth discrepancy between
branches provides a structurally consistent signal for dynamic
identification across diverse scene conditions, remaining
discriminative even when dynamic content is sparse.

\begin{table}[htpb]
\centering
\vspace{-1.0em}
\caption{\textbf{Dynamic Map Evaluation by Scene Dynamic Ratio.} 
We extend the evaluation in Tab.~6 by stratifying sequences into three 
groups based on their mean ground-truth dynamic ratio: Low ($\leq$10\%), 
Medium (10--30\%), and High ($>$30\%). The disc metric exceeds 1 across 
all groups, indicating that the depth discrepancy signal remains 
discriminative even when dynamic content is sparse.}
\label{tab:dynamic_ratio}
\scriptsize
\renewcommand{\arraystretch}{1.05}
\renewcommand{\tabcolsep}{3pt}
\resizebox{\linewidth}{!}{
\begin{tabular}{@{}lccc|cccc@{}}
\toprule
\textbf{Dynamic Ratio} & \textbf{Seq} & \textbf{Frames} & \textbf{Mean $\delta$} & \textbf{Mean disc} $\uparrow$ & \textbf{Mean AUC} $\uparrow$ & \textbf{Mean IoU} $\uparrow$ \\
\midrule
Low ($\leq$10\%)  & 66 & 4168 & 0.065 & 2.24 & 0.591 & 0.145 \\
Med (10--30\%)    & 45 & 2862 & 0.077 & 1.93 & 0.551 & 0.240 \\
High ($>$30\%)    & 20 &  985 & 0.153 & 1.36 & 0.532 & 0.409 \\
\bottomrule
\end{tabular}
}
\vspace{-2.0em}
\end{table}

\section*{F. Reset Metric Alignment}
Memory-based streaming models periodically reset their latent state to
mitigate state drift over long sequences~\cite{cut3r, ttt3r, point3r}.
While effective for stability, each reset alters the model's internal
scene representation, causing it to produce inconsistent metric scale
and pose for the same repeated frame when processed before and after
the reset. This discrepancy can manifest as scale mismatch between
segments, which may propagate as drift throughout subsequent frames.

Our reset metric alignment addresses this by using the repeated frame
as an anchor for cross-segment correction. Let $\mathbf{P}^{-}$ and
$\mathbf{P}^{+}$ denote the point clouds predicted from the repeated
frame before and after reset, respectively. Since both are predicted from the same input frame at identical pixel 
positions, correspondences are directly available without extra matching step. To focus the alignment on reliable scene structure,
we use the pixel-level staticness map computed at the final pre-reset
frame as per-point confidence weights, upweighting stable background
regions and downweighting potentially dynamic content. A Sim(3)
transformation is then estimated via confidence-weighted SVD,
recovering the scale, rotation, and translation discrepancy introduced
by the reset. The estimated transformation is applied to all subsequent
poses and accumulated point clouds in the new segment, reducing metric
inconsistency.

\begin{figure}[hptb]
 \vspace{-1.0em}
    \centering
    \includegraphics[width=0.7\linewidth]{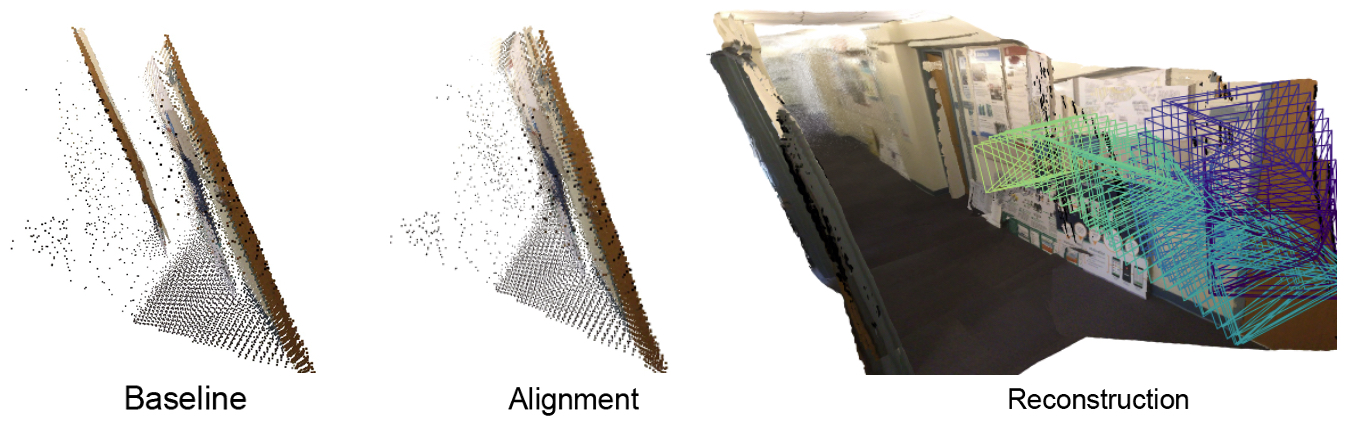}
\caption{
\textbf{Effect of Reset Metric Alignment.}
 Baseline (\emph{left}) produces fragmented point clouds across reset boundaries. Our reset metric alignment
restores segment consistency (\emph{middle}), yielding coherent
full-sequence reconstruction (\emph{right}).
}
    \label{fig:alignment}
    \vspace{-2.0em}
\end{figure}

\section*{G. State-Aware Smoothing Analysis}
We provide additional analysis of state-aware smoothing introduced in 
\cref{ssec:smoothing}, which stabilizes trajectory estimation by adaptively 
filtering inter-frame displacements. Specifically, we evaluate the contribution of each signal 
component, reporting Mean ATE and Mean RPE$_t$ averaged across 
Sintel, TUM-dynamics, and ScanNet under the same Sim(3) alignment protocol as in 
\cref{tab:camera_pose}. As state-aware smoothing primarily targets translational displacements, 
RPE$_r$ shows negligible variation across settings and is not reported.

As shown in \cref{tab:smoothing_ablation}, the smoothing coefficient $\beta_t \in [0,1]$ balances reliance on the 
current displacement prediction against accumulated history. Fixed-$\beta$ variants reduce RPE$_t$ over the 
unsmoothed baseline. However, uniform filtering cannot distinguish stable frames 
from uncertain ones, yielding limited improvement in ATE relative to the full 
adaptive scheme. Using $a_t$ alone yields ATE comparable to fixed-$\beta$ 
baselines. Trajectory acceleration alone cannot distinguish genuine motion 
dynamics from prediction noise, making $a_t$ an insufficiently selective signal. 
Using $\mathrm{sc}_t$ alone achieves competitive RPE$_t$ but degrades ATE. 
State change magnitude tends to be large at transitional intervals such as warmup 
and reset boundaries, causing over-smoothing that introduces systematic 
displacement bias and propagates as long-range drift. The product $a_t \times 
\mathrm{sc}_t$ restricts strong smoothing to frames where both signals are 
elevated, identifying noisy predictions while preserving reliable 
motion estimates. 
\begin{table*}[htpb]
\centering
\vspace{-1.0em}
\caption{\textbf{Smoothing Signal Ablation.} 
We compare the full adaptive smoothing ($a_t \times \mathrm{sc}_t$) against 
fixed-coefficient baselines and individual signal components. 
The full scheme achieves the lowest ATE, as jointly 
conditioning on both signals avoids under-suppression from fixed filtering 
and drift from single-signal over-smoothing.
}
\label{tab:smoothing_ablation}
\scriptsize
\renewcommand{\arraystretch}{1.05}
\renewcommand{\tabcolsep}{2pt}
\begin{tabular*}{\linewidth}{@{\extracolsep{\fill}}
  l|c|c|c|c|c|c|c@{}}
\toprule
& \textbf{Baseline} & \textbf{Fix $\beta{=}0.3$} & \textbf{Fix $\beta{=}0.5$} & \textbf{Fix $\beta{=}0.8$} & \textbf{Only $a_t$} & \textbf{Only $\mathrm{sc}_t$ } & \textbf{Full $a_t \!\times\! \mathrm{sc}_t$} \\
\midrule
Mean ATE $\downarrow$      & 0.112 & 0.091 & 0.092 & 0.095 & 0.091 & 0.120 & 0.080 \\
Mean $RPE_t$ $\downarrow$  & 0.036 & 0.022 & 0.026 & 0.033 & 0.028 & 0.021 & 0.021 \\
\bottomrule
\end{tabular*}
\vspace{-1.0em}
\end{table*}

\section*{H. Component Ablation}
To complement the main-paper ablation in Tab.~5, which samples sequences 
randomly across datasets for each task, we evaluate each component on 
complete individual datasets: camera pose estimation on 
TUM-dynamics~\cite{sturm2012benchmark}, video depth on 
KITTI~\cite{geiger2013vision}, and 3D reconstruction on 
7-Scenes~\cite{shotton2013scene}. As shown in \cref{tab:ablation_full}, 
R yields the largest improvement in video depth, as suppressing dynamic 
interference during state updates directly reduces per-frame depth 
error. S produces the most substantial gain in camera pose accuracy, 
as adaptive smoothing attenuates pose noise accumulated over dynamic 
sequences. M provides a consistent benefit to 3D reconstruction by 
correcting scale misalignment at reset boundaries, leading to more 
globally coherent point clouds. The full model achieves the best 
results across all metrics, suggesting that R, M, and S address 
complementary aspects of streaming reconstruction.

\begin{table*}[htpb]
\vspace{-1.0em}
\centering
\caption{\textbf{Cross-Dataset Component Ablation.}
Each task is evaluated on a representative dataset.
R: dual-branch RayMap identification; M: reset metric alignment;
S: state-aware smoothing. Each component contributes gains across tasks, and the full model 
achieves the best results across metrics. }
\label{tab:ablation_full}
\scriptsize
\renewcommand{\arraystretch}{1.05}
\renewcommand{\tabcolsep}{4pt}
\begin{tabular*}{\linewidth}{@{\extracolsep{\fill}}
  l|cc|cc|cc@{}}
\toprule
& \multicolumn{2}{c|}{\textbf{Camera Pose (TUM-dyn)}}
& \multicolumn{2}{c|}{\textbf{Video Depth (KITTI)}}
& \multicolumn{2}{c}{\textbf{3D Recon (7-Scenes)}} \\
\textbf{Setting}
& ATE $\downarrow$ & RPE$_t$ $\downarrow$
& AbsRel $\downarrow$ & $\delta{<}1.25$ $\uparrow$
& Acc $\downarrow$ & Chamfer $\downarrow$ \\
\midrule
Base  & 0.033 & 0.011 & 0.120 & 88.3 & 0.041 & 0.029 \\
+R    & 0.032 & 0.010 & 0.102 & 92.2 & 0.032 & 0.026 \\
+R+M  & 0.031 & 0.008 & 0.100 & 92.5 & 0.028 & 0.025 \\
+R+S  & 0.020 & 0.006 & 0.101 & 92.3 & 0.026 & 0.025 \\
Full  & \textbf{0.019} & \textbf{0.005} & \textbf{0.099} & \textbf{92.7} & \textbf{0.023} & \textbf{0.024} \\
\bottomrule
\end{tabular*}
\vspace{-1.0em}
\end{table*}

\section*{I. LLM Usage}
Large language models were used to assist with language polishing
in the preparation of this manuscript. All technical content,
experimental design, and conclusions are the work of the authors.

\end{document}